\documentclass[lettersize,journal]{IEEEtran}
\usepackage{amsmath,amsfonts}
\usepackage{algorithmic}
\usepackage{algorithm}
\usepackage{array}
\usepackage[caption=false,font=normalsize,labelfont=sf,textfont=sf]{subfig}
\usepackage{textcomp}
\usepackage{stfloats}
\usepackage{url}
\usepackage{verbatim}
\usepackage{graphicx}
\usepackage{cite}
\usepackage{subcaption}
\usepackage{booktabs}
\usepackage{multirow}
\usepackage{xcolor}
\usepackage{bm}
\usepackage{adjustbox}
\newcommand{\etal}{\textit{et al.}}
\begin{document}

\title{Improving 3D Gaussian Splatting Compression by Scene-Adaptive Lattice Vector Quantization}

\author{Hao Xu, Xiaolin Wu,~\IEEEmembership{Life Fellow,~IEEE,} and Xi Zhang,~\IEEEmembership{Member,~IEEE}
        % <-this % stops a space
%\thanks{This paper was produced by the IEEE Publication Technology Group. They are in Piscataway, NJ.}% <-this % stops a space
%\thanks{Manuscript received April 19, 2021; revised August 16, 2021.}}
\thanks{This work was supported in part by an internal grant of Southwest Jiaotong University and in part by the National Natural Science Foundation of China under Grant 62301313.}
\thanks{This work is based on Chapter 4 of the Ph.D. thesis of Hao Xu~\cite{xu2025thesis}, with additional experiments and revisions for journal publication.}
\thanks{H.~Xu is with the Department of Electrical \& Computer Engineering, McMaster University, Hamilton, ON L8S 4L8, Canada (email: xu338@mcmaster.ca).}
\thanks{X.~Wu is with the School of Computing and Artificial Intelligence, Southwest Jiaotong University, Chengdu, China (Corresponding author, email: xwu510@gmail.com).}
%\thanks{X.~Zhang is with the ANGEL Lab, Nanyang Technological University, Singapore. (email: xi.zhang@ntu.edu.sg).}
\thanks{X.~Zhang is with the School of Computer Science and Technology, Tongji University, China. (email: xzhang9308@gmail.com).}
}
% The paper headers
\markboth{Journal of \LaTeX\ Class Files,~Vol.~14, No.~8, August~2021}%
{Shell \MakeLowercase{\textit{et al.}}: A Sample Article Using IEEEtran.cls for IEEE Journals}

\IEEEpubid{0000--0000/00\$00.00~\copyright~2021 IEEE}
% Remember, if you use this you must call \IEEEpubidadjcol in the second
% column for its text to clear the IEEEpubid mark.

\maketitle

\begin{abstract}
3D Gaussian Splatting (3DGS) is rapidly gaining popularity for its photorealistic rendering quality and real-time performance, but it generates massive amounts of data. Hence compressing 3DGS data is necessary for the cost effectiveness of 3DGS models. 
Recently, several anchor-based neural compression methods have been proposed, achieving good 3DGS compression performance. 
However, they all rely on uniform scalar quantization (USQ) due to its simplicity. 
A tantalizing question is whether more sophisticated quantizers can improve the current 3DGS compression methods with very little extra overhead and minimal change to the system. 
The answer is yes by replacing USQ with lattice vector quantization (LVQ). To better capture scene-specific characteristics, we optimize the lattice basis for each scene, improving LVQ’s adaptability and R-D efficiency. This scene-adaptive LVQ (SALVQ) strikes a balance between the R-D efficiency of vector quantization and the low complexity of USQ. 
SALVQ can be seamlessly integrated into existing 3DGS compression architectures, enhancing their R-D performance with minimal modifications and computational overhead. 
Moreover, by scaling the lattice basis vectors, SALVQ can dynamically adjust lattice density, enabling a single model to accommodate multiple bit rate targets. This flexibility eliminates the need to train separate models for different compression levels, significantly reducing training time and memory consumption. 
\end{abstract}


\section{Introduction}
\IEEEPARstart{N}{ovel} view synthesis has witnessed remarkable progress in recent years, with Neural Radiance Fields (NeRF)~\cite{mildenhall2021nerf} established as a significant milestone in the field. 
While NeRF substantially improves rendering quality compared to its predecessors~\cite{lombardi2019neural,sitzmann2019scene,mildenhall2019local}, it suffers from slow rendering speeds due to the high computational cost of querying density and radiance values along camera rays.
More recently, 3D Gaussian Splatting (3DGS)~\cite{kerbl20233d} has rapidly gained popularity as an alternative approach, offering both high-quality rendering results and real-time performance. However, 3DGS requires a large number of Gaussian primitives to accurately represent a 3D scene, and storing these Gaussian attributes incurs substantial memory consumption. 
This motivates the research on compression of 3DGS models.

As redundancies exist in 3DGS models, some Gaussian primitives may be noncritical and not all attributes of these primitives require uniformly high precision to maintain rendering quality. Pruning and quantization can be leveraged to improve memory efficiency without materially sacrificing rendering performance~\cite{papantonakis2024reducing,lee2024compact,fan2023lightgaussian,wang2024end, navaneet2024compgs, niedermayr2024compressed}. 

However, these methods are limited in removing spatial redundancy because they largely ignore the correlations between neighboring Gaussian primitives.
To rectify this shortcoming, Scaffold-GS introduces a hierarchical representation by using a sparse set of anchors to generate a dense set of so-called neural Gaussians~\cite{lu2024scaffold}. 
Each anchor is associated with a group of neural Gaussians whose positions are defined by learnable offsets. The attributes of these Gaussians (i.e. opacity, color, rotation, scale) are dynamically predicted based on the anchor features and the viewing direction. 
Scaffold-GS lays the foundation for subsequent compression methods~\cite{chen2024hac,zhan2025catdgs,liu2024hemgs,chen2025hac++,wang2024contextgs}, which employ quantization and context-adaptive arithmetic coding to reduce the memory footprint of anchor primitives.
By estimating the conditional probability of each symbol given its context, these models reduce statistical redundancy and improve coding efficiency, representing the current state-of-the-art (SOTA) performance in 3DGS compression.

\IEEEpubidadjcol

\begin{figure*}
    \centering
    \includegraphics[width=0.98\linewidth]{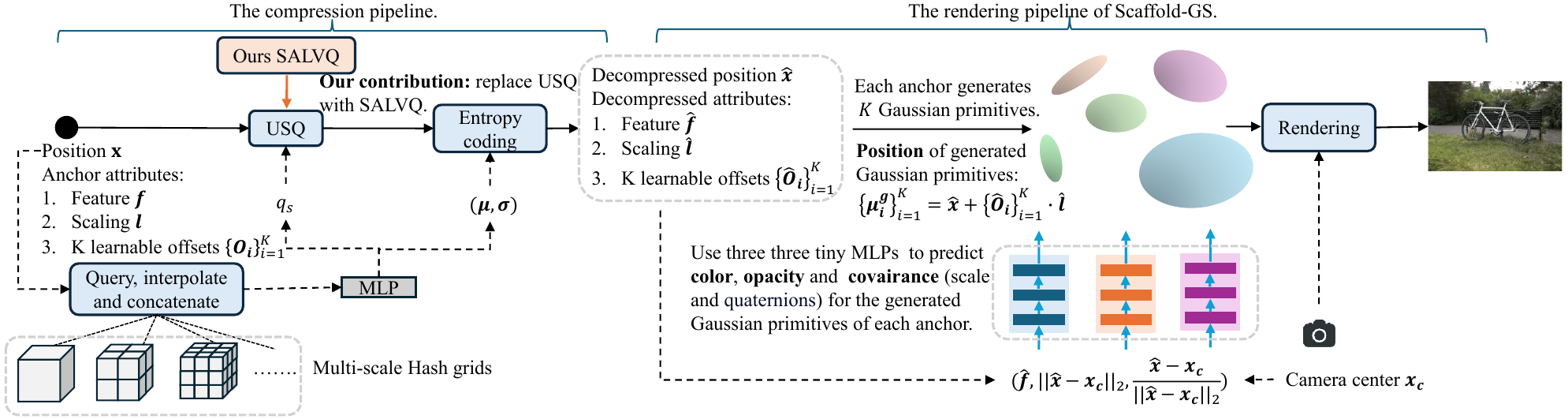}
    \caption{Illustrating the pipeline of anchor-based neural compression methods, using Hash-Assisted Context (HAC)~\cite{chen2024hac} as an example. Subsequent methods~\cite{wang2024contextgs, zhan2025catdgs, liu2024hemgs, chen2025hac++} follow this pipeline with different context models. Their superior R–D performance largely comes from learning a scene-specific Scaffold-GS representation and context model, whereas the quantizer remains a shared USQ. We go a step further by making the quantizer learnable and scene-adaptive so that its Voronoi partition better matches scene-specific statistics.}
    \label{fig:HAC}
\end{figure*}

A common drawback of anchor-based 3DGS compression methods is that they adopt uniform scalar quantization (USQ) to quantize anchor attributes. This naive scheme, although simplifies implementation, significantly compromises the rate-distortion (R-D) performance. 
Here a natural question arises: can we design more sophisticated quantizers that can improve the current SOTA methods with negligible overhead and minor changes to the system.  We answer the question affirmatively by replacing USQ with lattice vector quantization (LVQ). To better fit scene-specific feature distributions, we warp the lattice shape with respect to each scene, adapting LVQ to the scene structures and hence improving R-D efficiency. This scene-adaptive LVQ (SALVQ) strikes a balance between the coding efficiency of vector quantization (VQ) and the low complexity of USQ. SALVQ can be seamlessly embedded into existing 3DGS compression architectures with minimal modification, enhancing their R-D performance with almost no extra overhead. 
Moreover, by scaling the basis vectors of SALVQ, the proposed neural 3DGS neural compression model can dynamically adjust lattice density, enabling a single model to accommodate multiple target bit rates. This flexibility eliminates the need to train separate models for different compression levels, significantly reducing training time and memory consumption.

In summary, our main contribution is to use the novel scene-adaptive LVQ in 3DGS compression
to improve the storage and bandwidth economies of the 3DGS applications.  
The proposed SALVQ approach can facilitate the growing deployment of the 3DGS system,
owing to its following practical advantages:
\begin{itemize}
    \item Adaptability to specific scene statistics by warping LVQ cells to best fit the distribution of anchor features to be coded.

    \item The ability to support variable-rate 3DGS compression, offering flexibility in bitrate control while delivering high-quality reconstruction results.
       
    \item Compatibility with all neural compression architectures, meaning that the SALVQ method can be adopted with only minor modifications to existing systems.

\end{itemize}

\section{Related work}
\label{sec:review}

\subsection{3DGS and its compression}
\label{sec:review_3dgs}
The emerging 3DGS~\cite{kerbl20233d} represents scenes using a set of 3D Gaussian primitives and employs an efficient rasterization pipeline for rendering, achieving both high visual quality and real-time performance.
The memory footprint of a single Gaussian primitive is determined by storing its attributes, including position (3 floats), scale (3 floats), color (3 floats), rotation (4 floats for a quaternion), opacity (1 float), and Spherical Harmonics coefficients (45 floats), totaling 59 floats per primitive. After training, the number of Gaussian primitives can exceed millions, resulting in a memory footprint of several hundred MB or even several GB per scene. Such large memory consumption significantly restricts the deployment of 3DGS on memory-constrained devices and leads to substantial storage and transmission overhead. Thus, improving the memory efficiency of 3DGS has become a key research focus. Common methods include pruning redundant 3D Gaussians~\cite{girish2024eagles,xie2024mesongs,papantonakis2024reducing,wang2024end,fan2023lightgaussian,lee2024compact,navaneet2024compgs,chen2024hac,niemeyer2024radsplat,ali2024elmgs,hanson2024pup,liu2024efficientgs,lee2024safeguardgs,fang2024mini}, reducing the degree of SH coefficients~\cite{papantonakis2024reducing,wang2024end,fan2023lightgaussian}, vector quantization~\cite{wang2024end,fan2023lightgaussian,lee2024compact,niedermayr2024compressed,navaneet2024compgs} and attribute transform~\cite{girish2024eagles,xie2024mesongs}. Furthermore, there is a growing consensus that ideas from point cloud compression~\cite{zhang2014point,de2016compression,gu20193d,chou2019volumetric,sheng2022attribute,wang2022sparse,he2022density,xu2024fast} can inform and inspire the development of efficient 3DGS compression.

\subsection{Anchor-based GS and its compression}
The methods discussed in Sec.\ref{sec:review_3dgs} achieve only limited compression ratios because they overlook the inherent spatial organization among Gaussians, thus being classified as unstructured compression techniques. In contrast, structured compression methods explicitly exploit the spatial relationships and hierarchical organization of Gaussian representations for more efficient storage. The most prominent structured compression approach is Scaffold-GS\cite{lu2024scaffold}, which leverages anchors as structured reference points. Each anchor is characterized by its position $\mathbf{x}$, latent feature $\mathbf{f}$, scaling factor $\mathbf{l}$, and a set of $K$ learnable offsets $\{\mathbf{O}_i\}_{i=1}^{K}$. Each anchors generates $K$ Gaussian primitives whose positions depend on the anchor's scaling factor and offsets. The attributes of each Gaussian primitive (e.g., color, covariance, opacity) are then predicted from the anchor feature $\mathbf{f}$ and camera position $\mathbf{x}_c$ using a small MLP. The right panel of Fig.~\ref{fig:HAC} illustrates this anchor-based generation process for $K=5$.

Building upon the Scaffold-GS backbone~\cite{lu2024scaffold}, context modeling techniques~\cite{chen2024hac, wang2024contextgs, zhan2025catdgs, liu2024hemgs, chen2025hac++} have been incorporated into the entropy coding stage to reduce statistical redundancy among anchors, thereby achieving a more compact bitstream. 
Fig.~\ref{fig:HAC} illustrates the pipeline used by these methods, using hash-assisted context (HAC)~\cite{chen2024hac} as an example. As a prominent context modeling approach, HAC employs multi-scale hash grids as a hyperprior to effectively guide the entropy coder. Given the position of an anchor, HAC queries these hash grids to retrieve contextual features, which are subsequently processed by an MLP to predict the quantization scaling factor $q_s$, as well as the mean $\bm{\mu}$ and standard deviation $\bm{\sigma}$ of the Gaussian distribution for the anchor attributes. Building upon HAC, several more complex and powerful entropy models have been proposed. For example, HAC++~\cite{chen2025hac++} and CAT-3DGS~\cite{zhan2025catdgs} employ a channel-wise autoregressive model~\cite{minnen2020channel}, whereas ContextGS~\cite{wang2024contextgs} and HEMGS~\cite{liu2024hemgs} adopt spatial-wise autoregressive models~\cite{he2021checkerboard,minnen2018joint} to further enhance compression performance.

Although these methods achieve the current SOTA performance by learning a scene-adaptive Scaffold-GS representation and a matching scene-adaptive context model, they typically rely on USQ for simplicity. USQ's limited ability to accommodate diverse scene statistics leaves substantial room to improve R–D performance. This motivates us to propose a scene-adaptive quantizer that can be plugged into existing pipelines to further improve R–D performance with minimal modification to the backbone. 

\subsection{LVQ in neural image compression}
Given the advantages of LVQ that LVQ achieves a cost-effective balance between VQ and USQ, some researchers have explored the use of LVQ in neural image compression~\cite{zhang2023lvqac,lei2025approaching,cao2024entropy,zhang2024learning,Xu_2025_CVPR}. These approaches typically improve R-D performance with minimal modifications to existing network architectures. Generally, they employ a common LVQ across various images, with the lattice basis either being predefined~\cite{zhang2023lvqac,lei2025approaching} or optimized during training~\cite{zhang2024learning}. 
To enhance flexibility, Xu~\etal proposed a joint strategy for rate and domain adaptation by learning linear transforms of the lattice basis matrix~\cite{Xu_2025_CVPR}. Specifically, they achieve rate control by scaling the lattice basis vectors. For domain adaptation, they apply an invertible linear transforms to modulate the predefined lattice basis matrix, allowing it to better match the distinct characteristics of different image categories.

\begin{figure*}
    \centering
    \includegraphics[width=\linewidth]{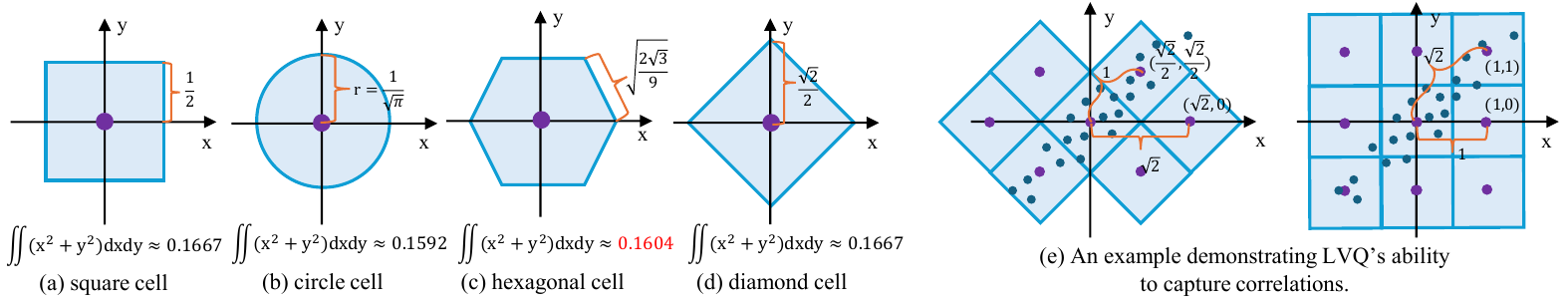}
    \caption{Illustration of the advantages of LVQ over USQ. For equal cell area, lattices whose Voronoi cells more closely approximate a circle yield a smaller mean-squared distance to the cell center than the canonical square (a–d). With equal-area diamond and square cells and correlated source components, diamond LVQ shortens the nearest-neighbor inter-codeword spacing along the principal direction, reducing the expected distortion; USQ leaves a larger spacing along the same principal direction (e).}
    \label{fig:lvq_idea}
\end{figure*}

\section{LVQ Preliminary}
\label{sec:lvq_pre_3dgs}
In preparation for the main technical development we introduce the basics of LVQ, including its concept and inherent coding advantages over scalar quantization.

Regular arrangements of points in vector space are called lattices~\cite{sayood2017introduction}. In $\mathbb{R}^n$, a lattice is formed by taking all integer linear combinations of a set of linearly independent basis vectors $\{\mathbf{b}_1, \mathbf{b}_2, \dots, \mathbf{b}_n\}$, denoted by
\begin{equation}
    \Lambda=\{\mathbf{z}|\mathbf{z}=\mathbf{Bu}=\sum_{i=1}^{n}u_i\mathbf{b}_i,~u_i\in\mathbb{Z}\}
    \label{eq:lattice_def}
\end{equation}
where $\mathbf{B}$ is the lattice basis matrix. A vector quantizer whose codewords are restricted to be lattice points is called lattice vector quantizer.

Thanks to the regular lattice structure, the nearest neighbor encoding of LVQ in $\mathbb{R}^n$ becomes very simple and can be performed in $O(n)$ time independent of the codebook size $L$, in contrast to the $O(nL)$ time of conventional VQ~\cite{conway1982fast}. 

Despite its regularity and low cost, LVQ still enjoys the coding benefit of VQ. Compared to USQ, LVQ has two inherent advantages: space-filling efficiency and the ability to capture correlation~\cite{gersho2012vector}. 
Fig.~\ref{fig:lvq_idea} (a-d) illustrates the space-filling advantage. Given the area/volume of a Voronoi cell, the optimal cell shape in terms of minimal mean squared quantization error is circle or hypersphere in high dimensions. However, tiling the space with spheres cannot avoid either overlaps or holes. LVQ addresses this issue by using a carefully designed lattice to partition a vector space into congruent convex Voronoi cells whose shape is as close to sphere as possible, achieving most efficient space covering. For some dimensions, optimal lattices are known to have Voronoi cells that best approximate spheres~\cite{conway2013sphere}. Examples include the hexagonal lattice ($A_2$) in $\mathbb{R}^{2}$, as well as the $\mathrm{E}_8$ lattice in $\mathbb{R}^{8}$ and Leech lattice in $\mathbb{R}^{24}$. 

Moreover, LVQ can capture the feature correlations in a way that USQ cannot. 
A representative example is the $D_n$ lattice, defined as $D_n=\{\mathbf{x}\in\mathbb{Z}^n:\sum_{i=1}^n x_i \text{ is even}\}$. In two dimensions, $D_2$ has a diamond-shaped Voronoi cell. For fair comparison with the square cell, Fig.~\ref{fig:lvq_idea}(d) uses a properly scaled version of $D_2$ whose Voronoi cell has the same area as the unit square cell.
Fig.~\ref{fig:lvq_idea} (e) depicts the case of correlation, where teal and purple points represent input vectors and their quantized codewords, respectively. In the diamond LVQ, the distance between two nearest quantizer codewords in the principal (diagonal) direction is 1; in contrast, the adjacent codeword distance in horizontal (non-principal) direction is $\sqrt{2}$.
By reducing quantization error along the principal direction, LVQ lowers the expected distortion and thus improves compression efficiency. 

\section{Method}
In this section, we first give an overview of LVQ design process. We then detail our method for optimizing LVQ on a per-scene basis and explain how it is integrated into an anchor-based neural 3DGS compression architecture. Finally, we develop a rate-control scheme that enables our LVQ-based 3DGS compression system to operate in variable-rate mode.
\subsection{Overview}
The anchor‑based 3DGS compression approach presents the current state of the art in the aspect of R–D performance. This is achieved by aggressive scene‑specific overfitting: a Scaffold‑GS scene representation and a conditional probability model for the entropy coding of anchor attributes are jointly trained per scene. As a result, these two components are highly tailored to each individual scene. Yet, as shown in Fig.~\ref{fig:HAC}, the quantizer remains a simple, scene‑agnostic uniform scalar quantizer (USQ); the only adjustable parameter is the step size. Because a USQ tiles the space with axis‑aligned hypercubes, its space‑filling efficiency is low, compromising compression performance. This naturally raises the question: can we also overfit a learnable LVQ that is jointly optimized with the Scaffold-GS and entropy model to further improve R-D performance? Our answer is emphatically yes.

To investigate this affirmative answer in practice, we begin by discussing the most direct and widely familiar approach: overfitting a separate codebook for each scene and applying VQ to the anchor attributes. In practice, however, conventional VQ is problematic. The encoding stage of VQ is non-differentiable. Although the soft-to-hard VQ~\cite{agustsson2017soft} and straight-through estimator (STE)~\cite{bengio2013estimating}-based approaches~\cite{van2017neural} can circumvent the non-differentiability issue, the former often struggles with low training stability, and the latter is prone to codebook collapse issue~\cite{takida2022sq}. Moreover, VQ inherently requires a nearest neighbor search during encoding, which increases computational cost. In addition, it needs to store a large per‑scene codebook, which increases the memory footprint and offsets the rate savings achieved by representing vectors with codebook indices. These drawbacks become more pronounced as the codebook size increases, making conventional VQ less practical for efficient 3DGS compression. 

In contrast, LVQ addresses several limitations of conventional VQ. LVQ generates the codebook implicitly from a learnable lattice basis matrix; only this compact basis needs to be stored, not a large explicit codebook. Because all code vectors are derived from a shared lattice structure, the number of learnable parameters is much smaller than in independently optimized codebooks, and this shared structure imposes geometric constraints that improve training stability. The lattice structure also admits fast quantization algorithms, reducing encoding cost. Together, these properties make LVQ a more practical and powerful replacement for USQ in 3DGS compression. While fixed lattice quantizers such as $D_n$ and $E_8$ are easy to deploy, their scene-agnostic nature prevents them from fully exploiting LVQ's potential. By contrast, jointly optimizing the lattice basis, Scaffold-GS, and the entropy model within a scene-adaptive LVQ, though more difficult, delivers greater returns. The following subsections detail the design of the proposed scene-adaptive LVQ.

\subsection{Learning scene-adaptive LVQ}
When learning SALVQ for 3DGS compression, one might consider leveraging two strategies previously explored in neural image compression. The first strategy is to directly optimize the lattice basis matrix while adding a regularization term to promote the orthogonality of the basis vectors~\cite{zhang2024learning}. The second strategy factorizes the lattice basis matrix $\mathbf{B}$ into the product of a learnable linear transform $\mathbf{A}$ and a predefined lattice basis matrix $\mathbf{G}$, i.e., $\mathbf{B}=\mathbf{A}\mathbf{G}$~\cite{Xu_2025_CVPR}. In this formulation, the linear transform $\mathbf{A}$ is constrained to be invertible and is parameterized as $\mathbf{A}=\mathbf{V}\mathbf{\Sigma}\mathbf{V^T}$, where $\mathbf{V}$ is a learnable rotation matrix derived from matrix exponential mapping and $\mathbf{\Sigma}$ is a learnable diagonal matrix. 
However, both strategies have notable drawbacks when directly applied to 3DGS compression. The first strategy may lead to training instability and can occasionally result in a non-invertible lattice basis matrix. On the other hand, the second strategy can only search for feasible LVQ solutions within a constrained region, which prevents the full potential of LVQ from being exploited. In some cases, the learned linear transform may degenerate into an identity transform when all diagonal elements approach one, leading to limited improvements in R-D performance. 

To overcome the above two shortcomings and enable LVQ search in a less constrained space, we propose a novel method to optimize SALVQ more effectively. Specifically, we parameterize the lattice basis matrix $\mathbf{B}$ using its singular value decomposition (SVD):
\begin{equation}
    \mathbf{B}=\mathbf{U\Sigma V^T}
    \label{eq:lvq_basis_svd}
\end{equation}
where $\mathbf{U}$ and $\mathbf{V}$ are learnable orthogonal linear transforms, and $\mathbf{\Sigma}$ is a learnable diagonal matrix with non-zero entries to ensure invertibility. We enforce the orthogonality of $\mathbf{U}$ and $\mathbf{V}$ using the orthogonal parametrization of PyTorch~\cite{paszke2019pytorch}, allowing flexible and stable optimization of $\mathbf{B}$. This SVD-based parameterization enables flexible and diverse lattice bases to be learned during training. In some special cases where all singular values approach one, the lattice basis matrix degenerates to $\mathbf{UV^T}$, which acts as a rotation matrix, resulting in the lattice Voronoi cell becoming a rotated hypercube in high-dimensional space. Notably, a suitably rotated hypercube can better capture correlations within feature vectors than the canonical hypercube that corresponds to the Voronoi cell of USQ, thereby reducing quantization error along principal directions.
In more general cases, this parameterization enables a rich set of feasible lattice basis matrix to be explored and optimized jointly with the 3DGS model. As a result, the system automatically discovers the SALVQ configuration that best suits the data, leading to improved R-D performance with negligible computational overhead. 

\subsection{Implementation details}
Because the learned lattice basis $\mathbf{B}$ is unconstrained, it will generally not coincide with highly structured lattices (e.g., $D_n$, $E_8$) that admit coset representations for fast exact quantization~\cite{conway1982fast}. To avoid computationally intensive nearest‑lattice‑point search under an arbitrary basis, we follow Zhang \etal~\cite{zhang2024learning} and adopt Babai's rounding technique (BRT)~\cite{babai1986lovasz}:
\begin{equation}
    q_l(\mathbf{y})=\mathbf{B}\lfloor \mathbf{B^{-1}y} \rceil
\end{equation}
where $\lfloor\cdot\rceil$ represents the rounding operation. BRT provides an efficient approximation to nearest‑lattice-point assignment under the basis $\mathbf{B}$, enabling fast quantization in practice.

Taking SALVQ for latent feature quantization as an example, before feeding the latent features $\mathbf{f}$ into the proposed SALVQ, we first perform mean removal to centralize the features. The shift vector used in this mean removal can be either spatially adaptive mean vectors $\boldsymbol{\mu}$ predicted by the hyperprior, or a global mean vector $\boldsymbol{\mu_g}$. When integrating SALVQ into a hyperprior-only architecture such as HAC~\cite{chen2024hac}, we adopt the first option. The centralized latent features, $\mathbf{f_c}=\mathbf{f}-\boldsymbol{\mu}$, are quantized using BRT. 
Specifically, we transform $\mathbf{f_c}$ into $\mathbf{f_t}=\mathbf{B^{-1}f_c}$ using the learned lattice basis matrix $\mathbf{B}$. 
Subsequently, we quantize $\mathbf{f_t}$ into $\mathbf{\hat{f}_t}$ through a rounding operation, which is approximated by adding uniform noise during training and replaced by hard quantization during inference.
The entropy coding of $\mathbf{\hat{f}_t}$ uses adaptive arithmetic coding, with the probability model for $\mathbf{\hat{f}_t}$ defined as: 
\begin{equation}
    p(\mathbf{\hat{f}_t})=\prod_{i}\Big[\Big(\mathcal{N}(0,\sigma_i^2)*\mathcal{U}(-\frac{q_s}{2},\frac{q_s}{2})\Big)(\hat{f}_t^{(i)})\Big]
\end{equation}
where $q_s$ is the quantization step size. In this framework, the hyperprior estimates spatially adaptive mean vectors for mean removal and predicts the variances of a zero-mean Gaussian distribution assumed for the quantized representation.
In the decompression stage, the $\mathbf{\hat{f}_t}$ is first decoded and then recovered to  $\mathbf{\hat{f}_c}=\mathbf{B}\mathbf{\hat{f}_t}$. Finally, the latent feature is reconstructed by adding the mean vectors $\boldsymbol{\mu}$ predicted by the hyperprior, i.e, $\mathbf{\hat{f}}=\mathbf{\hat{f}_c}+\boldsymbol{\mu}$. 

When integrating the proposed SALVQ approach into architectures with complex context models such as ContextGS~\cite{wang2024contextgs}, the mean vectors $\boldsymbol{\mu}$ for different groups are typically estimated sequentially. To avoid sequential quantization, we adopt a one-pass quantization strategy, ensuring that quantization is completed before entropy coding. To enable this one-pass quantization, we use a global mean vector $\boldsymbol{\mu_g}$ for mean removal. In this case, the assumption that $\mathbf{\hat{f}_t}$ follows a zero-mean Gaussian distribution becomes invalid, and the entropy models for $\mathbf{\hat{f}_t}$ remain the same as in the original architectures. Specifically, the entropy model for $\mathbf{\hat{f}_t}$ is a Gaussian distribution with a non-zero mean, and the parameters of these probability distributions are predicted using both the hyperprior and the causal context model.
\subsection{Design choices and extensions}
In anchor-based 3DGS compression, three types of anchor attributes require quantization: the latent features $\mathbf{f}$, the scaling factors $\mathbf{l}$, and the $K$ learnable offsets $\{\mathbf{O}_i\}_{i=1}^{K}$. In our default design, SALVQ is applied only to the latent features $\mathbf{f}$. This choice is motivated by two considerations. First, $\mathbf{f}$ accounts for the majority of the total bit budget and therefore has the largest impact on the overall compression performance. Second, since $\mathbf{f}$ is high-dimensional, it offers richer correlation structure and thus provides more room for LVQ-based quantization to improve over scalar quantization. 

Building upon this default design, we further explore two extensions of SALVQ. The first is to apply SALVQ to the anchor scaling factors $\mathbf{l}$. The second is a spatially adaptive variant, where anchors are partitioned into several groups according to their spatial locations and each group is assigned a dedicated SALVQ. 

By contrast, applying SALVQ to the learnable offsets $\{\mathbf{O}_i\}_{i=1}^{K}$ is less straightforward. Due to offset masking, the number of active offsets varies across anchors, resulting in non-uniform effective dimensionality of the offset vectors. However, vector quantization methods, including conventional VQ, LVQ, and the proposed SALVQ, require a fixed input dimensionality. In particular, SALVQ parameterizes the lattice basis matrix for a pre-defined input dimension, and therefore cannot be directly applied when the effective dimension varies across samples. For this reason, we do not consider applying SALVQ to the learnable offsets.

\subsection{Rate control scheme}
\label{sec:rate_control}
By replacing the USQ module in existing anchor-based 3DGS compression architectures with the proposed SALVQ, we enhance the system to achieve comparable or higher rendering quality with a smaller memory footprint. However, this system can only operate at a fixed rate. This is because the R-D trade-off is controlled by the Lagrange multiplier $\lambda$ in the loss function
\begin{equation}
\mathcal{L}=\mathcal{L}_{\mathrm{distortion}}+\lambda\mathcal{L}_{\mathrm{rate}}+\lambda_{\mathrm{reg}}\mathcal{L}_{\mathrm{reg}}
\end{equation}
where ${L}_{\mathrm{distortion}}$ denotes the rendering distortion, $\mathcal{L}_{\mathrm{rate}}$ corresponds to the rate, $\mathcal{L}_{\mathrm{reg}}$ represents other regularization terms controlled by $\lambda_{\mathrm{reg}}$. 
For a specific $\lambda$, the 3DGS model is optimized to achieve the corresponding rate target. For each choice of $\lambda$, the 3DGS model is trained to target a specific rate, and once training is complete, the model operates only at that rate. Supporting multiple R-D trade-offs requires training separate models with different $\lambda$ values, leading to computational and memory costs that grow linearly with the number of desired rates. This inefficiency highlights the importance of developing variable-rate compression methods that enable a single model to support multiple R-D trade-offs.

Understanding how the Lagrange multiplier $\lambda$ affects the R-D trade-off is essential for enabling variable-rate compression. 
A higher $\lambda$ penalizes the rate term more, reducing the rate at the cost of higher distortion. This occurs because a stronger penalty forces the anchor attributes to adopt a lower dynamic range, reducing the number of lattice points used. Decreasing $\lambda$ has the opposite effect, allowing a larger dynamic range and lower distortion. This mirrors neural image compression, where varying the quantization step size provides variable‑rate control~\cite{chen2020variable, cui2021asymmetric,tong2023qvrf, kamisli2024variable,Xu_2025_CVPR}; we adopt the same idea for 3DGS.
Motivated by this, we fix the anchor attributes' dynamic range and learn rate-specific lattice densities to control the R–D trade-off. Specifically, we introduce a gain vector $\mathbf{g} = [g_1, \cdots, g_M]$\footnote{The gain vector is learned separately for anchor latent features, scaling factors, and offsets.}, where $M$ is the number of target rates, to control the lattice density for each rate. The anchor attributes and the base quantization step $q_s$ are shared across all targets; target $i$ uses its gain $g_i$ to scale the quantization step from $q_s$ to $g_i q_s$. A larger gain leads to a coarser LVQ and a lower bitrate, enabling flexible rate control without retraining the model. To learn the gain vector, we associate each target with a Lagrange multiplier $\lambda_i$. At each training iteration, we randomly sample an index from $\{1, \cdots, M\}$, apply $g_i$ to modulate the step, and use $\lambda_i$ in the loss. After training, the anchor attributes and $q_s$ remain fixed; at inference one simply selects a learned gain $g_i$ to meet the desired bitrate.

This flexible rate control scheme is applicable to both SALVQ and USQ, noting that USQ can be considered as a special case of SALVQ where the identity matrix is used as the lattice basis matrix. When using this rate control scheme with either SALVQ or USQ, the distortion at each target rate is primarily determined by the quantization error. Under comparable bitrates, SALVQ typically achieves lower quantization error compared to USQ, resulting in lower distortion. Therefore, combining this rate control strategy with our proposed SALVQ leads to better overall performance.

\section{Experiment}
In this section, we describe the experimental setup; present the gains of SALVQ over USQ along with its memory and compute overhead; compare SALVQ-based 3DGS compression methods to earlier 3DGS compression baselines; assess visual quality differences between USQ- and SALVQ-based methods; evaluate both quantizers under variable-rate and progressive compression; and conclude with limitations and directions for future work.
\subsection{Experimental setting}
\subsubsection{Baselines} Our experiments are mainly based on three baselines: the earlier HAC~\cite{chen2024hac}, the more recent HAC++~\cite{chen2025hac++} and ContextGS~\cite{wang2024contextgs}. We also assess SALVQ on PCGS~\cite{chen2025pcgs}, a progressive 3DGS compression baseline. To evaluate the effectiveness of the proposed SALVQ approach, we replace the USQ module in these architectures with the proposed SALVQ, while keeping all other components unchanged. These baselines cover different design choices in the use of context models, including hyperprior-only, channel-wise autoregressive, and spatial autoregressive models, respectively. For reproducibility, we evaluate only baselines with stable, publicly available code. Accordingly, we exclude HEMGS~\cite{liu2024hemgs} (no released code) and CAT-3DGS~\cite{zhan2025catdgs} (issues in the current public release that prevent a reliable evaluation). Notably, both CAT-3DGS and HAC++ employ channel-wise autoregressive entropy models; our extensive results on HAC++ therefore already characterize SALVQ's behavior for this design family. We will include these additional baselines once stable, accessible implementations are available.

\subsubsection{SALVQ variants} We explore four variants of SALVQ, including one default setup and three extensions. 
\begin{itemize}
    \item \textbf{SALVQ} (the default configuration) applies SALVQ exclusively to the anchor latent feature $\mathbf{f}$. The full feature vector is treated as a single unit for vector quantization, and all anchors share a common SALVQ instance. Unless otherwise specified, ``SALVQ'' in the following experiments refers to this setup.
    \item \textbf{FS-SALVQ} extends this by applying SALVQ to both the anchor latent feature and the anchor scaling factors. 
    \item \textbf{Sp-SALVQ} (spatially adaptive SALVQ) partitions anchors into a small number of spatial groups and quantizes each group using its own dedicated SALVQ. We apply Sp-SALVQ to the ContextGS architecture as an extension of ContextGS + SALVQ, where a single global SALVQ is shared across all spatial locations. In other words, Sp-SALVQ replaces the shared quantizer with group-specific SALVQ quantizers, while keeping all other modules unchanged. The spatial grouping follows the same partition strategy as in ContextGS. Note that Sp-SALVQ is applied only to the anchor latent features, not to the scaling factors.
\end{itemize}

\subsubsection{Dataset} We evaluate the R-D performance on three commonly used large-scale real-scene datasets: Mip-NeRF360~\cite{barron2022mip}, Tanks\&Temples~\cite{knapitsch2017tanks}, and DeepBlending~\cite{hedman2018deep}. Notably, we assess all nine scenes from the Mip-NeRF360 dataset~\cite{barron2022mip}. These diverse datasets provide a comprehensive evaluation of the proposed SALVQ approach. 
\subsubsection{Distortion metrics} Following the evaluation settings commonly used in previous works, we evaluate rendering distortion using PSNR, SSIM~\cite{wang2004image}, and LPIPS~\cite{zhang2018unreasonable}. 
\subsubsection{Rate metrics} We use the size of the encoded bitstream (in MB) as the rate metric. To quantitatively evaluate the improvements introduced by the proposed SALVQ approach, we adopt the BD-rate~\cite{bjontegaard2001calculation} to measure the average reduction in memory footprint at a fixed distortion level. A negative BD-rate indicates that the compression system achieves a smaller memory footprint at the same quality compared to the baseline, with its absolute value reflecting the magnitude of the performance gain.

\subsubsection{Training details} Following the official settings, we train each scene for 30k iterations for HAC~\cite{chen2024hac}, HAC++~\cite{chen2025hac++}, and ContextGS~\cite{wang2024contextgs}, and for 40k iterations for PCGS~\cite{chen2025pcgs}.
The Lagrange multiplier $\lambda$ is varied over the set $\{0.002, 0.004, 0.008, 0.015, 0.025\}$ to evaluate compression performance across a wider range. 
For other hyperparameters, we follow the same settings as HAC~\cite{chen2024hac}, HAC++~\cite{chen2025hac++}, ContextGS~\cite{wang2024contextgs} and PCGS~\cite{chen2025pcgs}.

\begin{figure*}[ht]
\centering
\subfloat[HAC on Mip-NeRF360]{\includegraphics[width=2.2in]{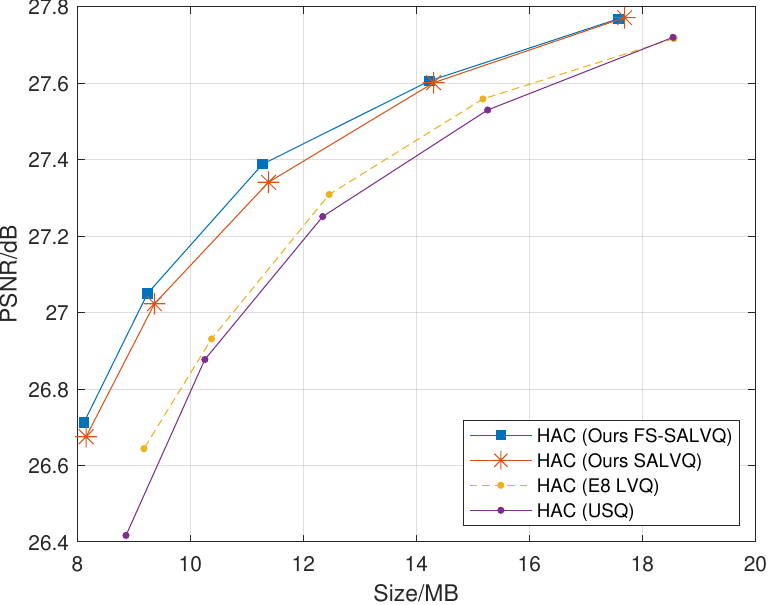}%
}
\hfil
\subfloat[HAC on Tank\&Temples]{\includegraphics[width=2.2in]{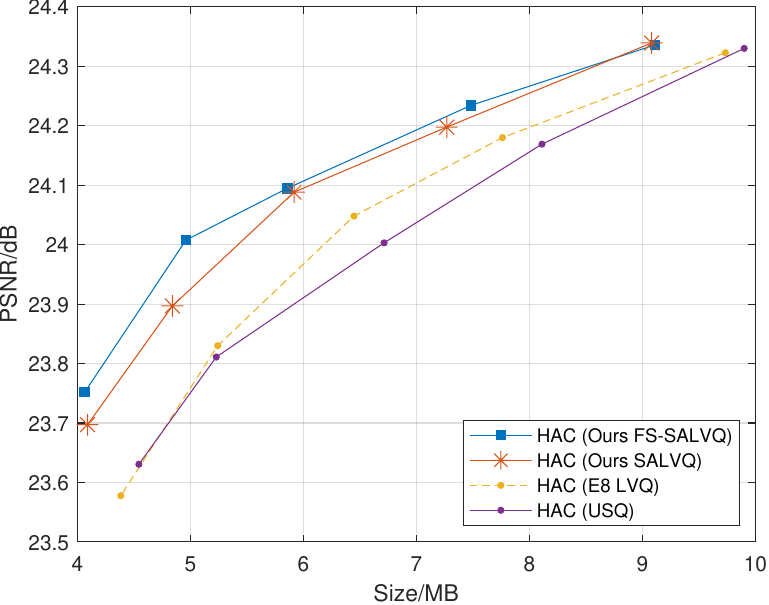}%
}
\hfil
\subfloat[HAC on DeepBlending]{\includegraphics[width=2.2in]{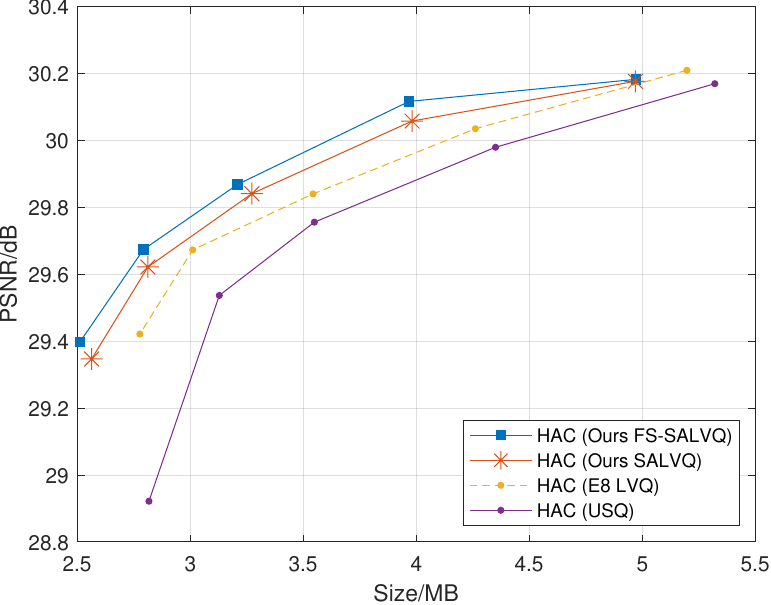}%
}
\\
\subfloat[ContextGS on Mip-NeRF360]{\includegraphics[width=2.2in]{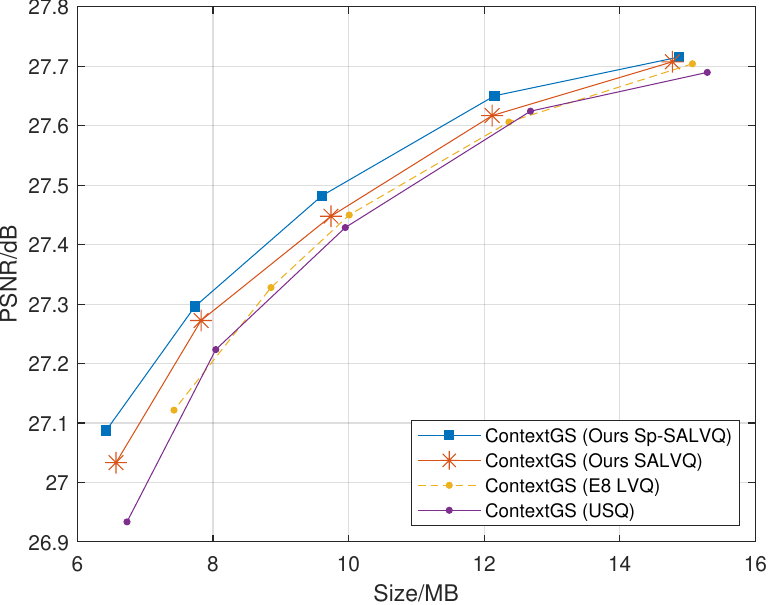}%
}
\hfil
\subfloat[ContextGS on Tank\&Temples]{\includegraphics[width=2.2in]{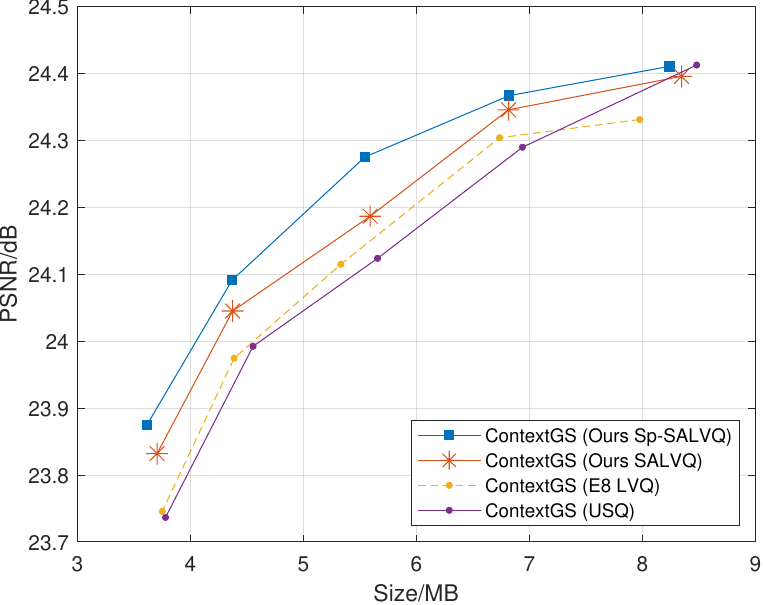}%
}
\hfil
\subfloat[ContextGS on DeepBlending]{\includegraphics[width=2.2in]{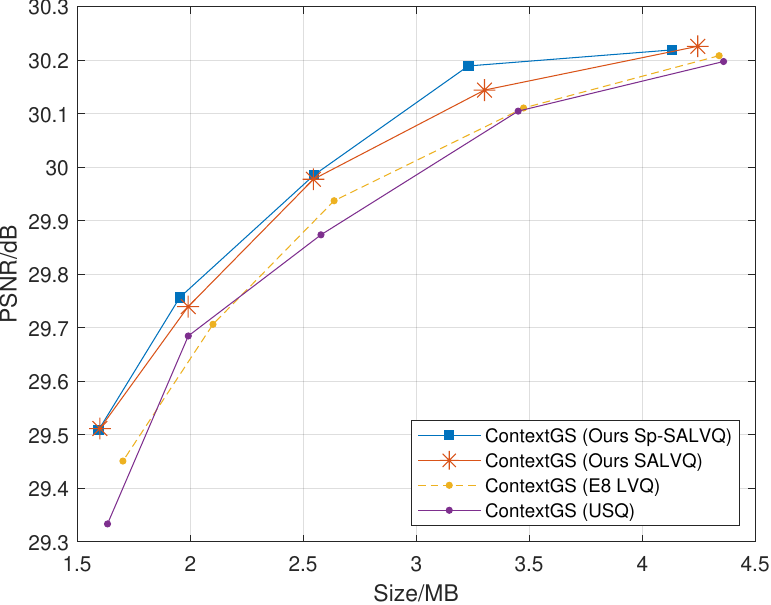}%
}
\\
\subfloat[HAC++ on Mip-NeRF360]{\includegraphics[width=2.2in]{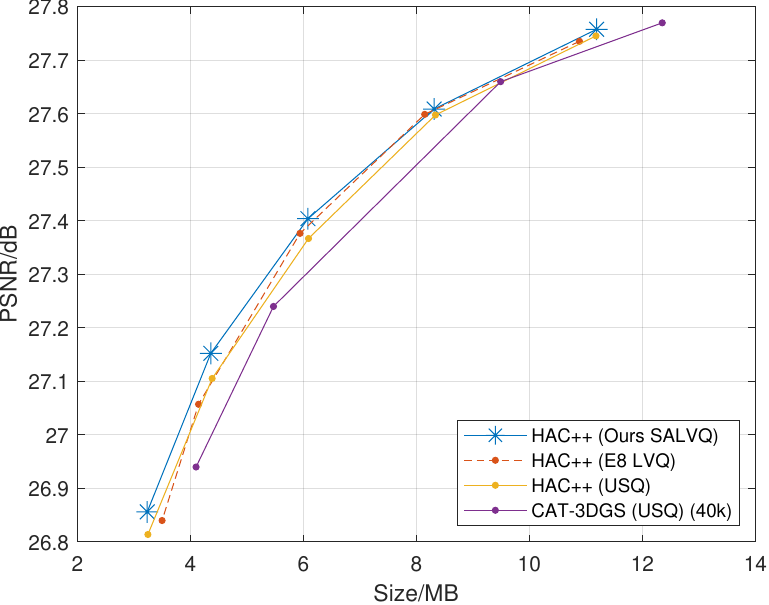}%
}
\hfil
\subfloat[HAC++ on Tank\&Temples]{\includegraphics[width=2.2in]{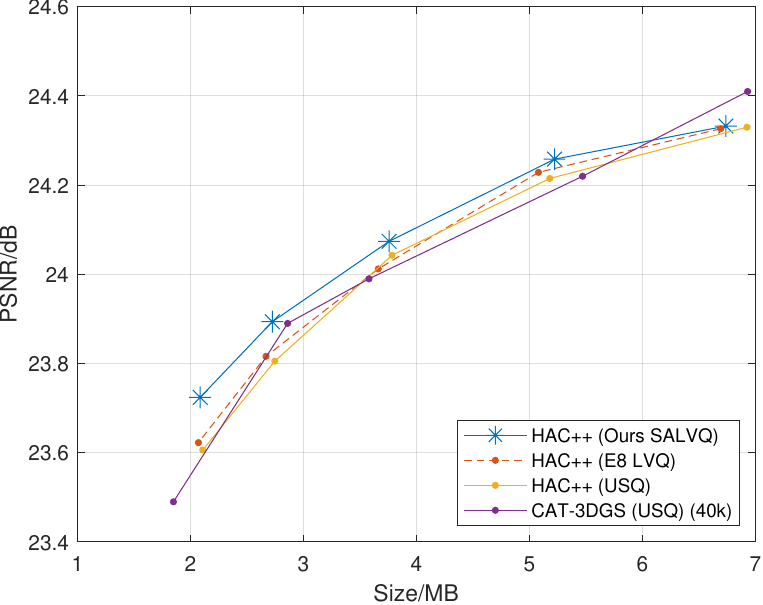}%
}
\hfil
\subfloat[HAC++ on DeepBlending]{\includegraphics[width=2.2in]{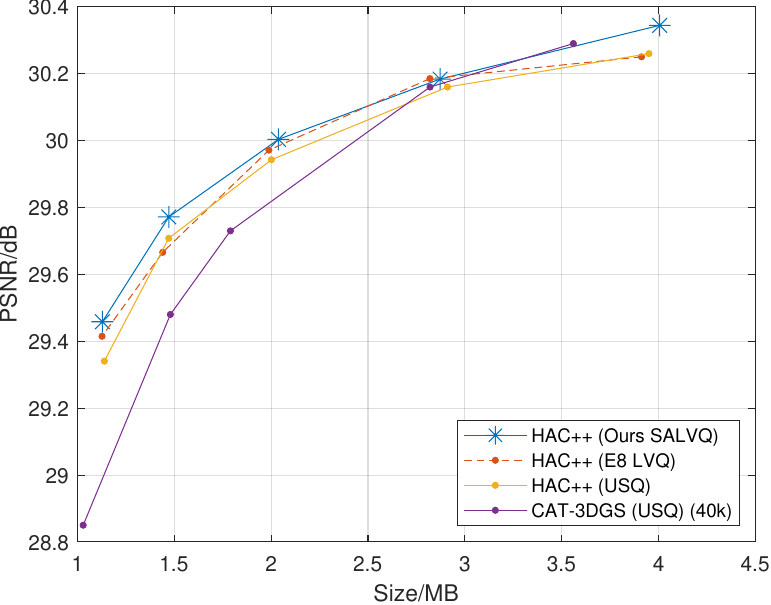}%
}
\caption{R–D curves of HAC~\cite{chen2024hac}, ContextGS~\cite{wang2024contextgs} and HAC++~\cite{chen2025hac++} under different quantizers.}
\label{fig:RD_curves}
\end{figure*}

\begin{table*}[!t]
\centering
\caption{BD-rate of LVQ relative to USQ on HAC~\cite{chen2024hac},  HAC++~\cite{chen2025hac++} and ContextGS~\cite{wang2024contextgs} across three datasets. Lower is better; negative denotes improved compression efficiency. \textbf{Bold} numbers highlight the best results.}
\footnotesize
\setlength{\tabcolsep}{3pt}
%\begin{adjustbox}{width=\linewidth}
\begin{tabular}{lcccccccccc}
        \toprule
         &\multicolumn{3}{c}{HAC~\cite{chen2024hac}}&&\multicolumn{2}{c}{HAC++~\cite{chen2025hac++}}&&\multicolumn{3}{c}{ContextGS~\cite{wang2024contextgs}}\\
         \cline{2-4}\cline{6-7}\cline{9-11}
         &$E_8$ LVQ&SALVQ&FS-SALVQ&&$E_8$ LVQ&SALVQ&&$E_8$ LVQ&SALVQ&Sp-SALVQ\\
         \midrule
         Mip-NeRF360~\cite{barron2022mip}&-2.04\%&{-13.48\%}&\textbf{-15.80\%} &&-1.52\%&\textbf{-4.55\%}& &-0.63\%&{-5.71\%}&\textbf{-9.77\%}\\
         Tank\&Temples~\cite{knapitsch2017tanks}&-4.42\%&{-16.16\%}&\textbf{-20.85\%}&&-2.35\%&\textbf{-8.95\%}&&-2.12\%&{-8.69\%}&\textbf{-15.60\%}\\
         DeepBlending~\cite{hedman2018deep}&-8.12\%&{-13.44\%}&\textbf{-17.09\%}&&-2.78\%&\textbf{-7.75\%}&&-0.93\%&{-9.75\%}&\textbf{-12.14\%}\\
         \bottomrule
\end{tabular}
%\end{adjustbox}
\label{tab:lvq_bd_rate}
\end{table*}

\subsection{Performance gain from SALVQ over USQ}
As shown in Fig.~\ref{fig:RD_curves}, the proposed SALVQ consistently improves the R-D performance over the USQ baseline across all architectures on all three datasets. Tab.~\ref{tab:lvq_bd_rate} further quantifies these gains: at the same distortion level, replacing USQ with SALVQ reduces the memory footprint, with average savings ranging from 4.55\% to 16.16\% across different architecture--dataset combinations. In addition, the two enhanced SALVQ extensions, namely FS-SALVQ and Sp-SALVQ, further improve upon the default SALVQ and achieve larger gains over the USQ baseline.

To highlight the benefit of scene-adaptive LVQ over a fixed lattice quantizer (such as one built with the $E_8$ lattice), we form a reference baseline by reducing the latent dimension from 50 to 48 and replacing USQ with a product $E_8$ lattice quantizer. As shown in Fig.~\ref{fig:RD_curves} and Tab.~\ref{tab:lvq_bd_rate}, our SALVQ consistently outperforms this fixed-lattice baseline. 
This is primarily because a fixed lattice quantizer is designed under the assumption of a uniform, scene-invariant latent distribution, which often does not align with the actual distribution of the latent features in each specific scene. Unlike generalizable compression, 3DGS compression is scene-specific; learning the lattice basis per scene can better align the quantizer with the true source statistics and significantly improve R-D performance.

We report per-scene results of pairing our SALVQ approach with existing 3DGS compression architectures, evaluated across multiple fidelity metrics (PSNR, SSIM~\cite{wang2004image}, and LPIPS~\cite{zhang2018unreasonable}) on three datasets. Detailed results are provided in the supplementary material.
\subsection{Comparison with CAT-3DGS}
Applying the proposed SALVQ to CAT-3DGS to directly evaluate its gain is unfortunately not feasible, because the current public release of CAT-3DGS has implementation issues that prevent reliable evaluation. To still provide a meaningful comparison with this strong concurrent baseline, we include an additional comparison based on the R-D results reported in the CAT-3DGS paper. Specifically, we insert the reported CAT-3DGS R-D curves into the third row of Fig.~\ref{fig:RD_curves} and compute the BD-rate of HAC++ (USQ) and HAC++ (Ours SALVQ) relative to CAT-3DGS. We use PSNR as the distortion metric for this comparison, since its computation is consistent across methods and thus provides a fair basis for BD-rate evaluation from the reported R-D curves. As shown in Fig.~\ref{fig:RD_curves} and Tab.~\ref{tab:cat_3dgs}, even though CAT-3DGS uses 40k training iterations while all HAC++ variants are trained for only 30k iterations, the basic HAC++ (USQ) baseline already outperforms the reported CAT-3DGS results, and combining HAC++ with SALVQ further substantially enlarges the BD-rate gain relative to CAT-3DGS.

\begin{table}[]
    \centering
    \caption{BD-rate of HAC++ variants using two different quantizers, relative to CAT-3DGS.}
    %\begin{adjustbox}{width=\linewidth}
    \begin{tabular}{lcc}
    \toprule
         &HAC++ (USQ) &HAC++ (Ours SALVQ)\\
        \midrule
         Mip-NeRF360~\cite{barron2022mip}&-4.41\% &-8.66\%\\
         Tank\&Temples~\cite{knapitsch2017tanks}&-3.10\% &-7.78\%\\
         DeepBlending~\cite{hedman2018deep}&-9.61\% &-15.08\%\\
         \bottomrule
    \end{tabular}
    %\end{adjustbox}
    \label{tab:cat_3dgs}
\end{table}

\subsection{Discussion of SALVQ gains across different architectures}
Fig.~\ref{fig:RD_curves} and Tab.~\ref{tab:lvq_bd_rate} show a clear trend: pairing SALVQ with a simpler backbone such as HAC~\cite{chen2024hac} yields larger gains than pairing it with more complex backbones such as HAC++~\cite{chen2025hac++} and ContextGS~\cite{wang2024contextgs}. This behavior is as expected. More generally, entropy models and quantizers improve compression performance by addressing partially overlapping aspects of redundancy in the latent representation, although they do so in different ways. A stronger entropy model can already capture more correlation and structure in the latent, which naturally reduces the room left for a more sophisticated quantizer to provide additional benefit. This is precisely why the gain of SALVQ is larger on HAC than on HAC++ or ContextGS.

A smaller marginal gain with stronger entropy models does not imply that SALVQ loses its value. SALVQ still achieves consistent and measurable BD-rate reductions on stronger backbones across datasets, showing that its benefit does not disappear as entropy models improve. Moreover, given the low-complexity nature of SALVQ, it remains beneficial to retain these gains rather than discard them. Even when entropy models become stronger through more complex designs, it is difficult for them to fully exploit all redundancy in the latent representation on their own. In this sense, SALVQ should be viewed not as a replacement for entropy modeling, but as a useful complementary tool that further improves coding efficiency at limited additional cost.

From a practical deployment perspective, a compression method should be evaluated not only by its R-D performance, but also by its computational cost and coding latency. In learned compression systems, the entropy model, especially when autoregressive, is often a major contributor to runtime overhead and latency. Therefore, the large gains of SALVQ on simpler entropy models are particularly meaningful in practice: they suggest that one can achieve a strong rate-distortion-complexity trade-off without relying on increasingly complex entropy modeling. For low-complexity 3DGS compression scenarios, combining SALVQ with a simple entropy model is therefore a particularly attractive design choice.

\begin{table}[!t]
\centering
\caption{Comparison of rendering FPS between USQ and SALVQ variants.}
\footnotesize
\setlength{\tabcolsep}{3pt}
\begin{adjustbox}{width=\linewidth}
\begin{tabular}{lcccccccccc}
        \toprule
         &\multicolumn{3}{c}{HAC~\cite{chen2024hac}}&&\multicolumn{2}{c}{HAC++~\cite{chen2025hac++}}&&\multicolumn{3}{c}{ContextGS~\cite{wang2024contextgs}}\\
         \cline{2-4}\cline{6-7}\cline{9-11}
         &USQ&SALVQ&FS-SALVQ&&USQ&SALVQ&&USQ&SALVQ&Sp-SALVQ\\
         \midrule
         Mip-NeRF360~\cite{barron2022mip}&131 &136 &136 &&128& 143& &106 &104&102\\
         Tank\&Temples~\cite{knapitsch2017tanks}&146 &139 &149&&158 &164&&130&130&129\\
         DeepBlending~\cite{hedman2018deep}&179 &173 &199&&169&182&& 161&169&194\\
         \bottomrule
\end{tabular}
\end{adjustbox}
\label{tab:lvq_fps}
\end{table}

\begin{table}[!t]
\centering
\caption{Comparison of encoding time (in seconds) between USQ and SALVQ variants.}
\footnotesize
\setlength{\tabcolsep}{3pt}
\begin{adjustbox}{width=\linewidth}
\begin{tabular}{lcccccccccc}
        \toprule
         &\multicolumn{3}{c}{HAC~\cite{chen2024hac}}&&\multicolumn{2}{c}{HAC++~\cite{chen2025hac++}}&&\multicolumn{3}{c}{ContextGS~\cite{wang2024contextgs}}\\
         \cline{2-4}\cline{6-7}\cline{9-11}
         &USQ&SALVQ&FS-SALVQ&&USQ&SALVQ&&USQ&SALVQ&Sp-SALVQ\\
         \midrule
         Mip-NeRF360~\cite{barron2022mip}&4.20& 4.56&4.71 &&8.48& 8.37& &32.73& 31.93&34.47\\
         Tank\&Temples~\cite{knapitsch2017tanks}&2.51 &2.80&2.79&&5.63 & 5.30&&29.37 & 31.64&31.87\\
         DeepBlending~\cite{hedman2018deep}&1.33 &1.49&1.45&&2.98 &2.91&&16.78 & 12.59&11.73\\
         \bottomrule
\end{tabular}
\end{adjustbox}
\label{tab:lvq_encoding_time}
\end{table}

\begin{table}[!t]
\centering
\caption{Comparison of decoding time (in seconds) between USQ and SALVQ variants.}
\footnotesize
\setlength{\tabcolsep}{3pt}
\begin{adjustbox}{width=\linewidth}
\begin{tabular}{lccccccccccc}
        \toprule
         &\multicolumn{3}{c}{HAC~\cite{chen2024hac}}&&\multicolumn{2}{c}{HAC++~\cite{chen2025hac++}}&&\multicolumn{3}{c}{ContextGS~\cite{wang2024contextgs}}\\
         \cline{2-4}\cline{6-7}\cline{9-11}
         &USQ&SALVQ&FS-SALVQ&&USQ&SALVQ&&USQ&SALVQ&Sp-SALVQ\\
         \midrule
         Mip-NeRF360~\cite{barron2022mip}&10.05 & 10.76&10.85 &&13.86& 13.81& &33.32&31.84&33.74\\
         Tank\&Temples~\cite{knapitsch2017tanks}&5.77 &6.43&6.12&&8.81& 8.61&&29.05& 28.80&30.73\\
         DeepBlending~\cite{hedman2018deep}&2.85 &3.11&3.00&&4.41& 4.42&&16.63&12.49&11.37\\
         \bottomrule
\end{tabular}
\end{adjustbox}
\label{tab:lvq_decoding_time}
\end{table}

\begin{table}[!t]
\centering
\caption{Comparison of GPU vRAM consumption during training (in MB) between USQ and SALVQ variants.}
\footnotesize
\setlength{\tabcolsep}{3pt}
\begin{adjustbox}{width=\linewidth}
\begin{tabular}{lcccccccccc}
        \toprule
         &\multicolumn{3}{c}{HAC~\cite{chen2024hac}}&&\multicolumn{2}{c}{HAC++~\cite{chen2025hac++}}&&\multicolumn{3}{c}{ContextGS~\cite{wang2024contextgs}}\\
         \cline{2-4}\cline{6-7}\cline{9-11}
         &USQ&SALVQ&FS-SALVQ&&USQ&SALVQ&&USQ&SALVQ&Sp-SALVQ\\
         \midrule
         Mip-NeRF360~\cite{barron2022mip}&8228 &10197 &10462 &&9732&11654 & &11738 &12012&12854\\
         Tank\&Temples~\cite{knapitsch2017tanks}&4130 &5493 &6540&&4026 &6534&&5841&5958&6194\\
         DeepBlending~\cite{hedman2018deep}&5729 &8175 &8254&&6361&7681&& 6789&6884&7404\\
         \bottomrule
\end{tabular}
\end{adjustbox}
\label{tab:lvq_GPUvram}
\end{table}

\begin{table}[!t]
\centering
\caption{Comparison of training time (in seconds) between USQ and SALVQ variants.}
\footnotesize
\setlength{\tabcolsep}{3pt}
\begin{adjustbox}{width=\linewidth}
\begin{tabular}{lcccccccccc}
        \toprule
         &\multicolumn{3}{c}{HAC~\cite{chen2024hac}}&&\multicolumn{2}{c}{HAC++~\cite{chen2025hac++}}&&\multicolumn{3}{c}{ContextGS~\cite{wang2024contextgs}}\\
         \cline{2-4}\cline{6-7}\cline{9-11}
         &USQ&SALVQ&FS-SALVQ&&USQ&SALVQ&&USQ&SALVQ&Sp-SALVQ\\
         \midrule
         Mip-NeRF360~\cite{barron2022mip}&1949 & 2206&2265 &&2735& 2937& &3927 &4167&4427\\
         Tank\&Temples~\cite{knapitsch2017tanks}&1385 & 1701&1750&&1880 &2055&&2505&2639&2844\\
         DeepBlending~\cite{hedman2018deep}&1533 & 1766&1843&&1857&2060&&2320 &2375&2636\\
         \bottomrule
\end{tabular}
\end{adjustbox}
\label{tab:lvq_training_time}
\end{table}

\subsection{Memory and computational overhead}
Beyond the BD-rate gains, the cost of achieving the R–D gains is pivotal in persuading developers to adopt SALVQ rather than USQ. In this subsection, we evaluate the memory and compute costs introduced by SALVQ.

\paragraph{Memory overhead} Applying SALVQ to the anchor latent feature, which has a dimension of 50, introduces only $50^2 + 50^2 + 50=5050$ additional parameters to parameterize the adaptive lattice basis matrix $\mathbf{B}$ (see Eq.~(\ref{eq:lvq_basis_svd})). Storing these parameters in float32 incurs a memory overhead of merely 0.02 MB, which is negligible relative to the memory footprint of the bitstream. Further applying SALVQ to the anchor scaling factors, which have a dimension of 6, adds only $6^2+6^2+6=78$ extra parameters. Partitioning all anchors into a small number of subsets (e.g., 3) multiplies this parameter overhead by the same factor, yet the total overhead remains negligible.

\paragraph{Runtime overhead}
As shown in Tabs.~\ref{tab:lvq_fps}--\ref{tab:lvq_decoding_time}, SALVQ has little adverse effect on runtime performance compared with USQ. Owing to the inherently low complexity of LVQ, replacing USQ with SALVQ keeps the encoding and decoding times broadly comparable. Moreover, rendering is performed after decompression on the recovered 3DGS parameters, meaning that the quantizer is not involved during rendering and the rendering pipeline remains unchanged. As a result, the rendering FPS remains broadly comparable to that of the USQ-based baselines. The enhanced SALVQ variants, namely FS-SALVQ and Sp-SALVQ, also maintain similar runtime efficiency.

\paragraph{GPU vRAM overhead}
Tab.~\ref{tab:lvq_GPUvram} reports the peak GPU vRAM consumption during training. As shown in the table, SALVQ and its two enhanced variants incur a moderate increase in peak GPU vRAM consumption during training compared with the USQ-based baseline. However, this increase is manageable in practice and does not lead to out-of-memory issues.

\paragraph{Training-time overhead of SALVQ}
Tab.~\ref{tab:lvq_training_time} show that SALVQ introduces additional training time, with overhead ranging from 55 to 316 seconds. Given that the USQ-based baselines already require 1385--3927 seconds for training, this additional cost is moderate in practice. More importantly, in most cases, this overhead is incurred only once during offline 3DGS asset preparation for each scene, rather than repeatedly during deployment. Since decoding and rendering are executed many times after deployment, a modest increase in one-time training cost in exchange for improved rate-distortion performance while preserving inference efficiency is a reasonable trade-off.

\paragraph{Training-time overhead of FS-SALVQ and Sp-SALVQ}
As shown in Tab.~\ref{tab:lvq_training_time}, further applying SALVQ to the anchor scaling factors leads to only a slight increase in training time compared with the default SALVQ setup. In contrast, decomposing anchors into several subsets and learning a dedicated SALVQ for each subset introduces some additional training-time overhead compared with using a single shared SALVQ. However, this additional cost is still moderate in practice, amounting to only about 200 to 300 seconds.

\begin{figure*}[!t]
\centering
\subfloat[]{\includegraphics[width=2.8in]{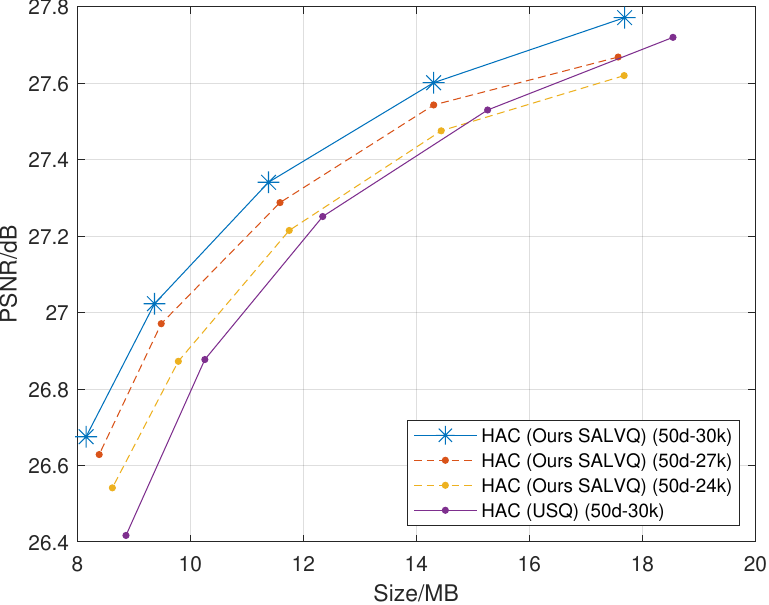}%
}
\hfil
\subfloat[]{\includegraphics[width=2.8in]{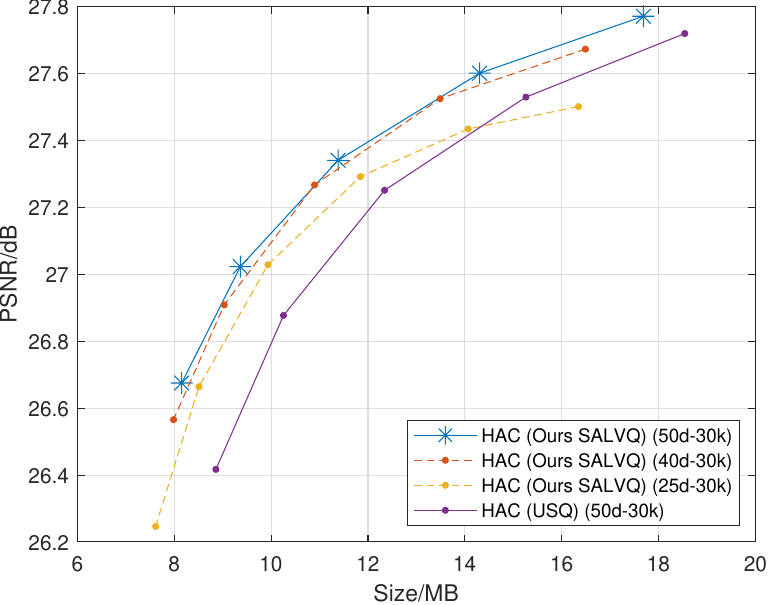}%
}
\caption{(a) R-D performance of HAC + SALVQ on Mip-NeRF 360 under different training iterations (30k, 27k, and 24k), with the anchor feature dimensionality fixed at 50. (b) R-D performance of HAC + SALVQ on Mip-NeRF 360 under different latent dimensionalities (50d, 40d, and 25d), with all models trained for 30k iterations.}
\label{fig:training_iteration_latent_dimension}
\end{figure*}

\paragraph{Coding gains of SALVQ under different training budgets}
To further examine the trade-off between training-time overhead and coding gain, we conduct an additional experiment on HAC using the Mip-NeRF 360 dataset. We use this dataset because it contains nine scenes spanning both indoor and outdoor environments, making it a representative benchmark. We compare SALVQ and USQ under different training budgets while fixing the anchor latent feature dimensionality to 50 in all cases. As reported in Tab.~\ref{tab:training_time_bd_rate} and Fig.~\ref{fig:training_iteration_latent_dimension}(a), when both methods are trained for 30k iterations, SALVQ increases the training time from 1949\,s to 2206\,s (about 13\%) while achieving a 13.48\% BD-rate reduction. When the training budget of SALVQ is reduced, it still retains considerable gains: with 27k iterations, its training time (1911\,s) becomes comparable to that of the USQ baseline trained for 30k iterations, while still achieving an 8.97\% BD-rate reduction; with a further reduced schedule of 24k iterations, SALVQ uses only about 83\% of the training time of the USQ baseline trained for 30k iterations and still delivers a 2.83\% BD-rate reduction. These results show that SALVQ maintains clear coding gains even under comparable or smaller training budgets.

In summary, the proposed SALVQ approach is low-complexity and cost-effective: it delivers significant R-D performance improvements with little additional computation or parameters, making it a strong candidate for integration as a basic module in 3DGS compression systems.

\begin{table}[]
    \centering
    \caption{Training-time overhead and corresponding BD-rate gains of replacing USQ with SALVQ in HAC under different training schedules. The SALVQ training budget is controlled by reducing the number of training iterations from 30k to 27k and 24k, and all BD-rate values are computed relative to HAC (USQ) (50d-30k).}
    \begin{tabular}{ccc}
    \toprule
    &Training time& BD-rate\\
    \midrule
        HAC (USQ) (50d-30k) &1949 &0 \\
        \hline
        HAC (Our SALVQ) (50d-30k) & 2206 &-13.48\%\\
        HAC (Our SALVQ) (50d-27k)  & 1911 &-8.97\%\\
        HAC (Our SALVQ) (50d-24k)  & 1621 &-2.83\%\\
        \bottomrule
    \end{tabular}
    \label{tab:training_time_bd_rate}
\end{table}

\begin{table*}[]
    \centering
    \caption{Training and inference efficiency of SALVQ with reduced latent dimensionality in HAC, together with BD-rate gains relative to the 50d USQ baseline.}
    \begin{tabular}{ccccccc}
    \toprule
    &Training time&GPU vRAM&Rendering FPS&Encoding time&Decoding time& BD-rate\\
    \midrule
        HAC (USQ) (50d-30k) &1949&8228 &131 &4.20&10.05&0\\
        \hline
        HAC (Our SALVQ) (50d-30k) & 2206 &10197 &136&4.56&10.76&-13.48\%\\
        HAC (Our SALVQ) (40d-30k)  & 2012& 9694&141&4.44&10.69&-12.27\%\\
        HAC (Our SALVQ) (25d-30k)  & 1887&9045 &145&4.39&10.48&-7.94\%\\
        \bottomrule
    \end{tabular}
    %\caption{Training and inference efficiency of SALVQ with reduced latent dimensionality in HAC, together with BD-rate gains relative to the 50d USQ baseline.}
    \label{tab:reduce_latent_dim}
\end{table*}

\begin{table*}[ht]
\centering
\footnotesize
\setlength{\tabcolsep}{6pt}
\renewcommand{\arraystretch}{1.2}
\caption{Comparison between SALVQ-based compression systems and other 3DGS compression methods, including 3DGS and Scaffold-GS, for reference. \textbf{Bold} numbers and \underline{underline} highlight the best and the second-best results, respectively. The size values are measured in megabytes (MB).}
%\vspace{-2mm}
\begin{tabular}{l|cccc|cccc|cccc}
\hline
%\rowcolor[HTML]{FFFFFF} 
\textbf{Datasets}  & \multicolumn{4}{c|}{\textbf{Mip-NeRF360~\cite{barron2022mip}}} & \multicolumn{4}{c|}{\textbf{Tank\&Temples~\cite{knapitsch2017tanks}}} & \multicolumn{4}{c}{\textbf{DeepBlending~\cite{hedman2018deep}}}  \\ \hline
%\rowcolor[HTML]{FFFFFF} 
\textbf{Methods} & \textbf{psnr$\uparrow$} & \textbf{ssim$\uparrow$} & \textbf{lpips$\downarrow$} & \textbf{size$\downarrow$}  & \textbf{psnr$\uparrow$} & \textbf{ssim$\uparrow$} & \textbf{lpips$\downarrow$} & \textbf{size$\downarrow$}  & \textbf{psnr$\uparrow$} & \textbf{ssim$\uparrow$} & \textbf{lpips$\downarrow$} & \textbf{size$\downarrow$}  \\ \hline
3DGS~\cite{kerbl20233d} & 27.46 & \textbf{0.812} & \textbf{0.222} & 750.9 & 23.69 & 0.844 & \underline{0.178} & 431.0 & 29.42 & 0.899 & \textbf{0.247} & 663.9  \\ 
Scaffold-GS~\cite{lu2024scaffold} & 27.50 & 0.806 & 0.252 & 253.9 & 23.96 & \textbf{0.853} & \textbf{0.177} & 86.50 & \textbf{30.21} & \underline{0.906} & 0.254 & 66.00 \\ 
\hline

Compact3DGS~\cite{lee2024compact} & 27.08 & 0.798 & 0.247 & 48.80 & 23.32 & 0.831 & 0.201 & 39.43 & 29.79 & 0.901 & 0.258 & 43.21  \\ 
Compressed3D~\cite{niedermayr2024compressed} & 26.98 & 0.801 & 0.238 & 28.80 & 23.32 & 0.832 & 0.194 & 17.28 & 29.38 & 0.898 & 0.253 & 25.30  \\ 
EAGLES~\cite{girish2024eagles} & 27.14 & \underline{0.809} & 0.231 & 58.91 & 23.28 & 0.835 & 0.203 & 28.99 & 29.72 & \underline{0.906} & \underline{0.249} & 52.34 \\ 
LightGaussian~\cite{fan2023lightgaussian} & 27.00 & 0.799 & 0.249 & 44.54 & 22.83 & 0.822 & 0.242 & 22.43 & 27.01 & 0.872 & 0.308 & 33.94  \\ 
SOG \textit{et al.}~\cite{morgenstern2024compact} & 26.56 & 0.791 & 0.241 & 16.70 & 23.15 & 0.828 & 0.198 & 9.30 & 29.12 & 0.892 & 0.270 & 5.70  \\ 
Navaneet \textit{et al.}~\cite{navaneet2024compgs} & 27.12 & 0.806 & {0.240} & 19.33 & 23.44 & 0.838 & 0.198 & 12.50 & 29.90 & \textbf{0.907} & {0.251} & 13.50 \\ 
Reduced3DGS~\cite{papantonakis2024reducing} & 27.19 & 0.807 & \underline{0.230} & 29.54 & 23.57 & 0.840 & 0.188 & 14.00 & 29.63 & 0.902 & {0.249} & 18.00  \\ 
%\hline
RDOGaussian~\cite{wang2024end} & 27.05 & 0.802 & {0.239} & 23.46 & 23.34 & 0.835 & 0.195 & 12.03 & 29.63 & 0.902 & {0.252} & 18.00  \\ 
CompGS~\cite{liu2024compgs}&27.26&0.803&0.239&16.50&23.70&0.837&0.208&9.60&29.69&0.901&0.279&8.77\\
\hline
FCGS~\cite{chen2025fast} & 27.05 & 0.798 & 0.237 & 36.30 & 23.48 & 0.833 & 0.193 & 18.80 & 29.27 & 0.893 & 0.257 & 30.10 \\
\hline
HAC~\cite{chen2024hac}  & 27.53 &0.807 & 0.238 & 15.26 & {24.04} & 0.846 & {0.187} & 8.10 & 29.98 & 0.902 & 0.269 &  {4.35} 
\\ 
HAC~\cite{chen2024hac} + Ours SALVQ  & 27.60 &0.807 & 0.239 & 14.30 & 24.20 & 0.847 & 0.186 & 7.27 & 30.06 & 0.903 & 0.268 & 3.98 
\\ 
HAC~\cite{chen2024hac} + Ours FS-SALVQ  & 27.61 &0.807 & 0.239 & 14.23 & 24.23 & 0.848 & 0.185 & 7.48 & 30.12 & 0.903 & 0.268 & 3.96 
\\ 
\hline
Context-GS~\cite{wang2024contextgs} & \underline{27.62} & {0.808} & 0.237 & 12.68 & {24.20} & {0.852} & {0.184} & 7.05 & {30.11} & \textbf{0.907} & 0.258 & 3.45
\\
ContextGS~\cite{wang2024contextgs} + Ours SALVQ & \underline{27.62} & 0.808 & 0.238 & {12.12} & \underline{24.35} & \underline{0.852} & 0.184 & {6.81} & 30.14 & \textbf{0.907}& 0.266 & {3.30} 
\\
ContextGS~\cite{wang2024contextgs} + Ours Sp-SALVQ & \textbf{27.65} & 0.808 & 0.237 & {12.15} & \textbf{24.37} & {0.849} & 0.186 & {6.82} & \underline{30.19} & {0.904}& 0.268 & {3.23} 
\\
\hline
HAC++~\cite{chen2025hac++}  &  {27.60} & {0.803} & 0.253 & \underline{8.34} & {24.22} & 0.849 & {0.190} & \textbf{5.18} & {30.16} & \textbf{0.907} & 0.266 &  \underline{2.91}\\
HAC++~\cite{chen2025hac++} + Ours SALVQ& {27.61} & 0.803 & 0.252 & \textbf{8.31} & {24.26} & 0.849 & 0.190 & \underline{5.22} & {30.18} & \textbf{0.907} & 0.266 & \textbf{2.87} \\
\hline

\end{tabular}
\label{tab:main_results}
%\vspace{-4mm}
\end{table*}

\subsection{SALVQ under reduced latent feature dimensions}
To evaluate the effectiveness of SALVQ under reduced latent dimensionality, we conduct an experiment on HAC using the Mip-NeRF360 dataset, which contains nine scenes spanning both indoor and outdoor environments and thus serves as a representative benchmark. We reduce the latent dimensionality from 50 to 40 and 25 while keeping the training schedule fixed at 30k iterations. These reduced-dimensional SALVQ variants are compared against the original 50d USQ-based HAC baseline, with results reported in Tab.~\ref{tab:reduce_latent_dim} and Fig.~\ref{fig:training_iteration_latent_dimension}(b). The 40d SALVQ variant achieves a 12.27\% BD-rate reduction relative to the 50d USQ baseline, while also reducing training time and peak GPU vRAM consumption compared to the 50d SALVQ variant. Remarkably, even the 25d SALVQ variant still outperforms the 50d USQ baseline overall, with a 7.94\% BD-rate reduction. A closer inspection of the R-D curves, however, shows that the 25d SALVQ variant mainly surpasses the 50d USQ baseline in the low-bitrate regime, while showing a bottleneck at high bitrates. This suggests that SALVQ can still provide clear compression gains even with a compact low-dimensional latent representation, whereas more aggressive dimensionality reduction limits high-bitrate, high-fidelity reconstruction due to reduced feature capacity.

\begin{figure*}
    \centering
    \includegraphics[width=0.95\linewidth]{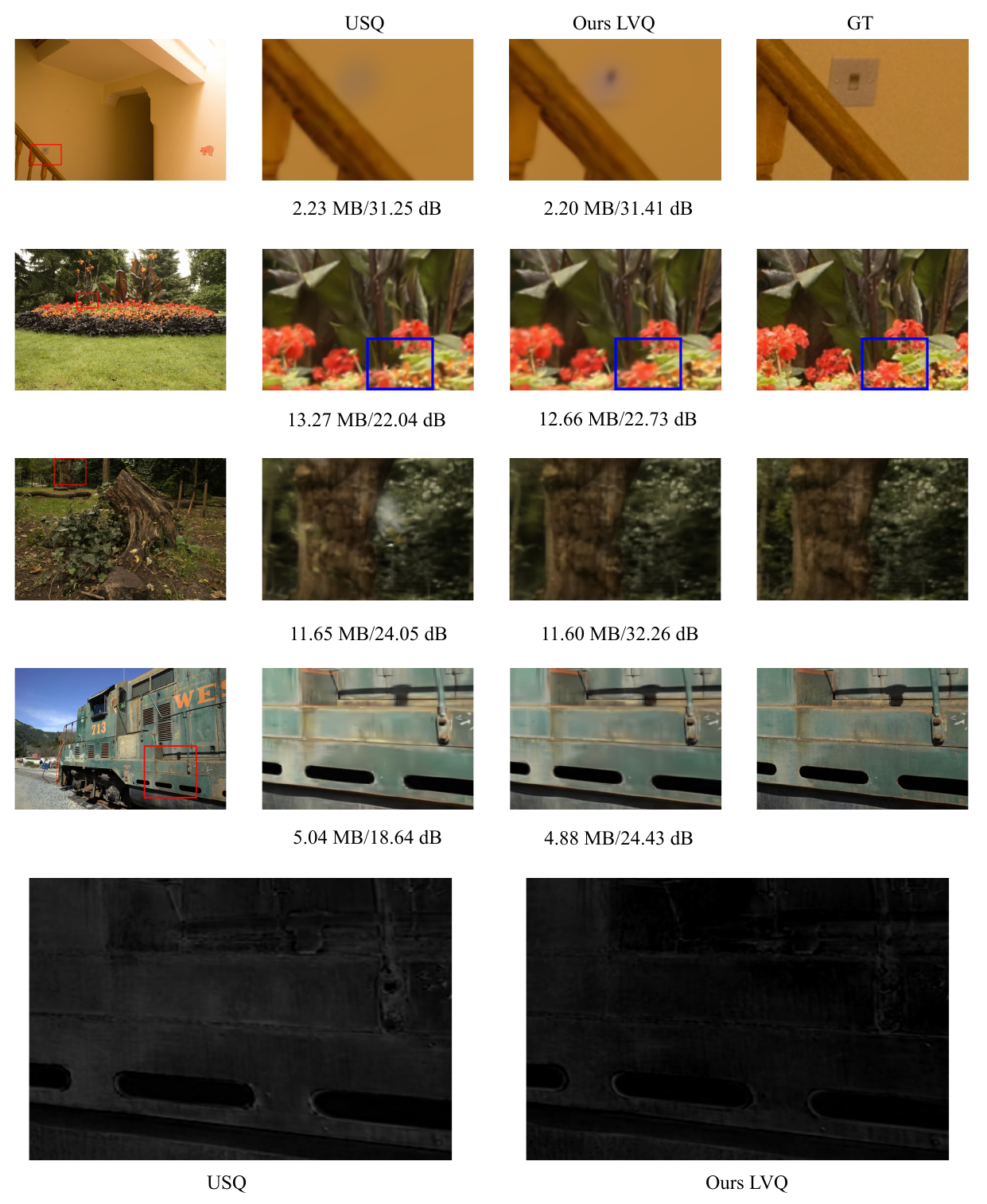}
    
    \caption{Visual comparison of different quantizers on four scenes: `playroom' (DeepBlending~\cite{hedman2018deep}), `flower' and `stump' (Mip-NeRF360~\cite{barron2022mip}), and `train' (Tanks and Temples~\cite{knapitsch2017tanks}), shown from the first to the fourth row, respectively. The last row shows the Y-channel residual maps for both quantization methods, providing a more intuitive visualization of the differences in brightness preservation on the `train' scene.}
    \label{fig:visual}
\end{figure*}

\begin{figure*}[ht]
\centering
\subfloat[HAC on Mip-NeRF360]{\includegraphics[width=2.2in]{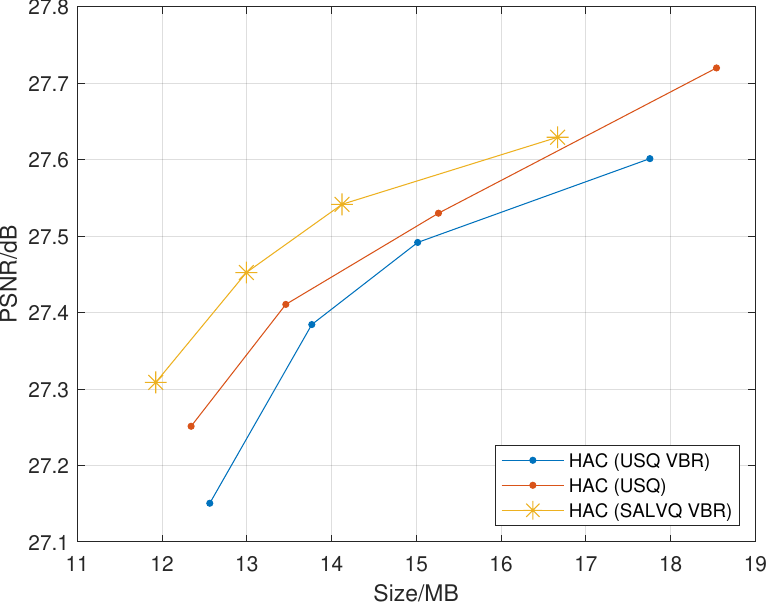}%
}
\hfil
\subfloat[HAC on Tank\&Temples]{\includegraphics[width=2.2in]{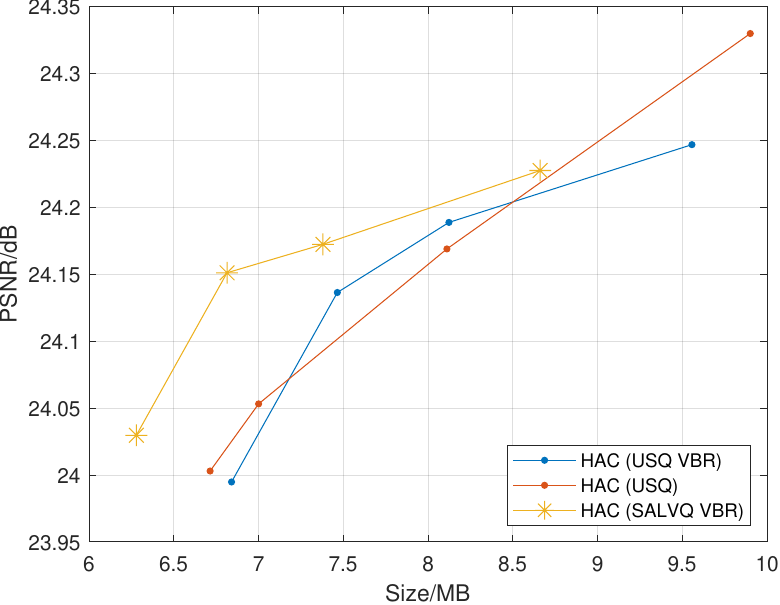}%
}
\hfil
\subfloat[HAC on DeepBlending]{\includegraphics[width=2.2in]{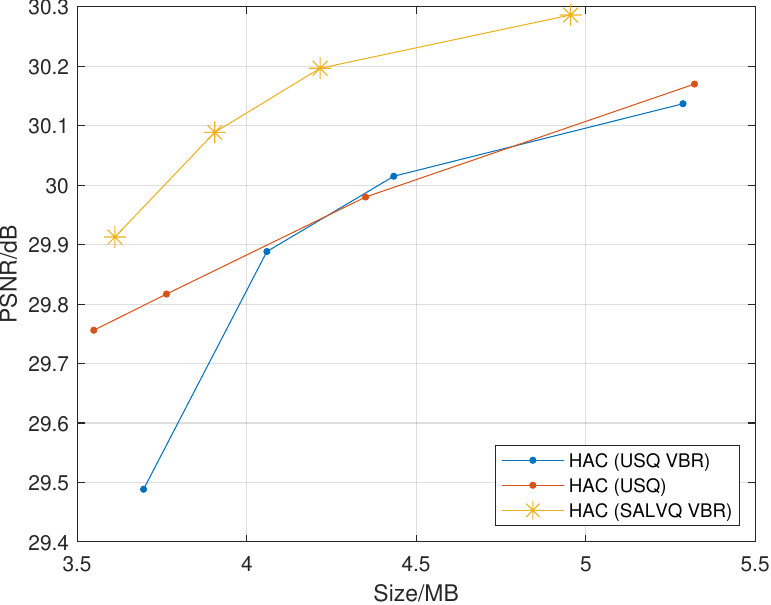}%
}
\\
\subfloat[ContextGS on Mip-NeRF360]{\includegraphics[width=2.2in]{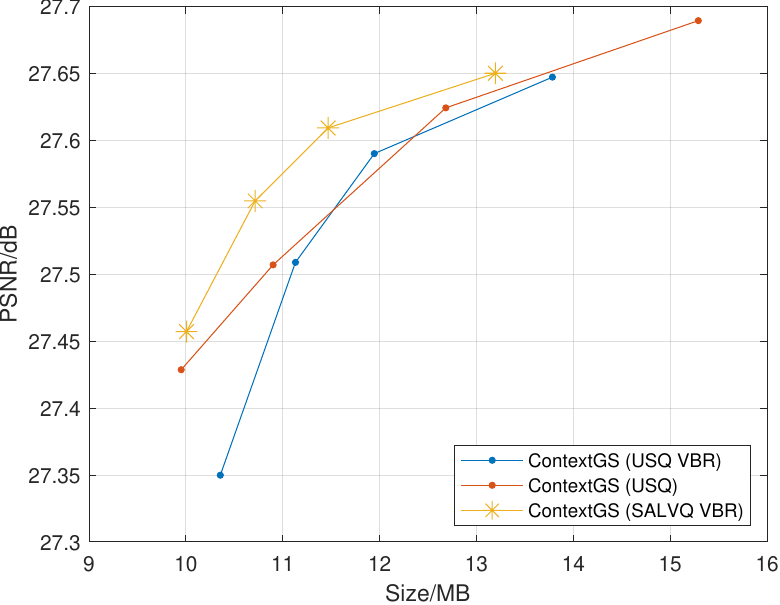}%
}
\hfil
\subfloat[ContextGS on Tank\&Temples]{\includegraphics[width=2.2in]{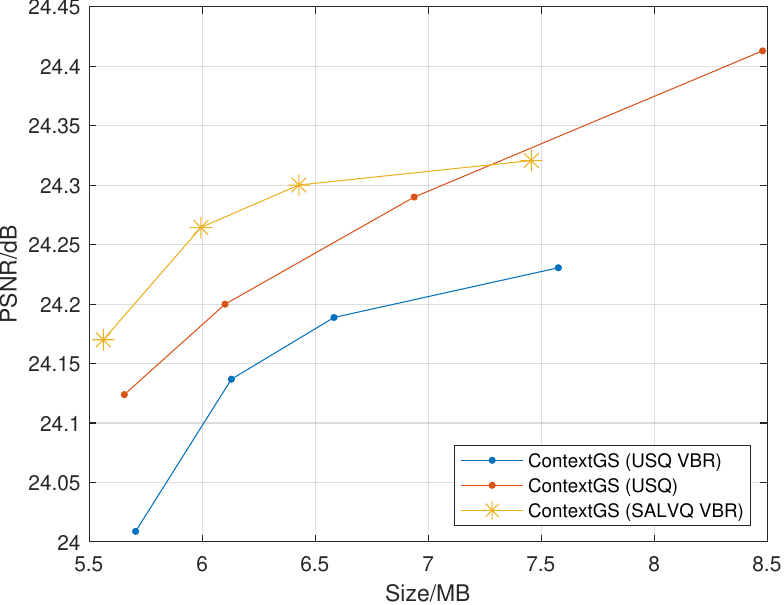}%
}
\hfil
\subfloat[ContextGS on DeepBlending]{\includegraphics[width=2.2in]{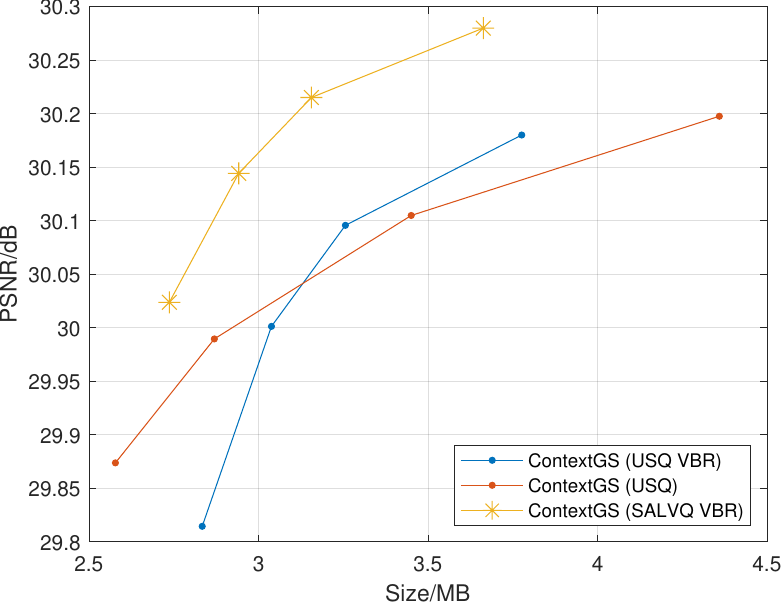}%
}
\\
\subfloat[HAC++ on Mip-NeRF360]{\includegraphics[width=2.2in]{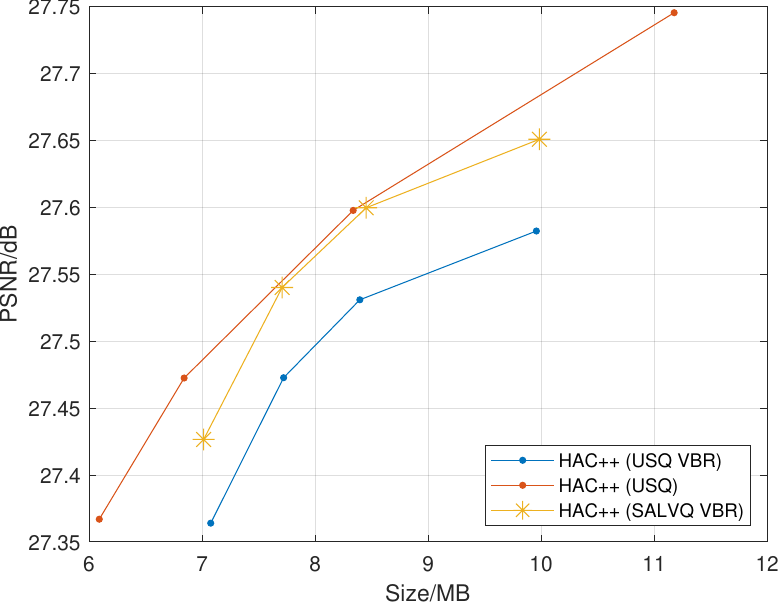}%
}
\hfil
\subfloat[HAC++ on Tank\&Temples]{\includegraphics[width=2.2in]{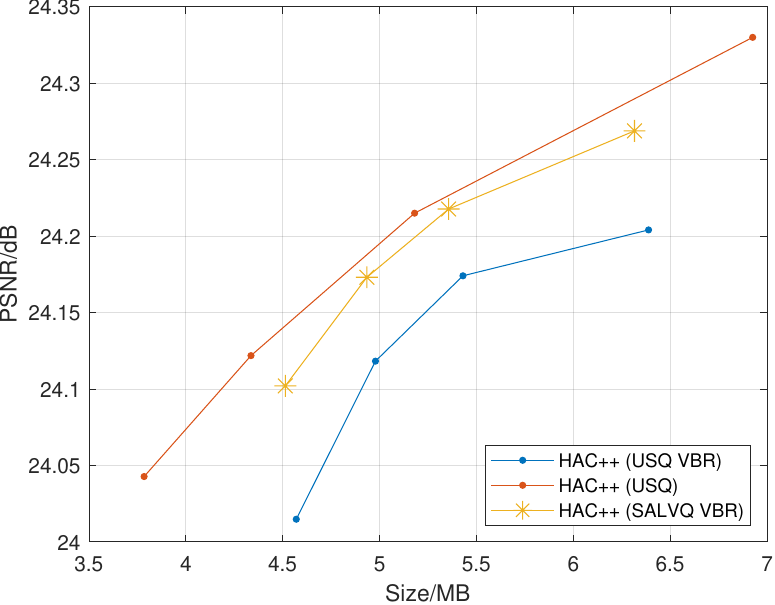}%
}
\hfil
\subfloat[HAC++ on DeepBlending]{\includegraphics[width=2.2in]{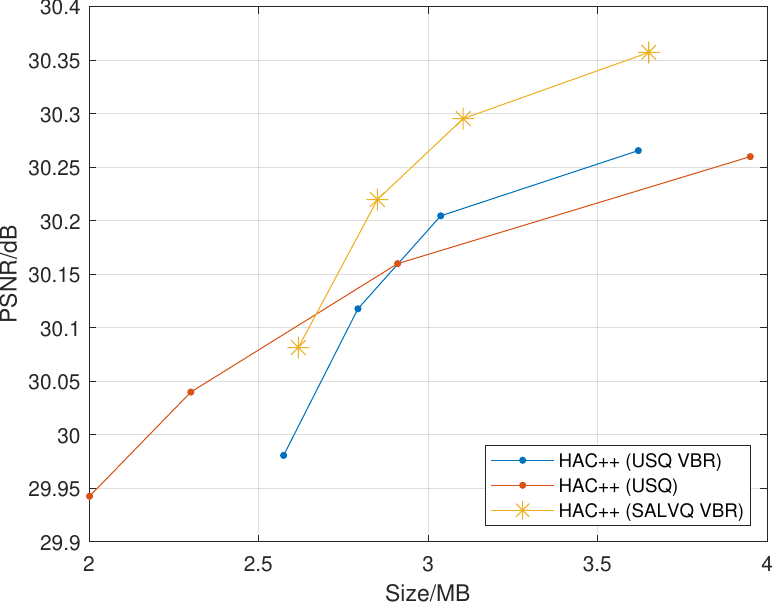}%
}

\caption{Single-rate vs. variable-rate compression performance on three architectures and datasets. Here, `VBR' indicates that the R-D curve corresponds to a variable rate compression method.}
\label{fig:RD_curves_vbr}
\end{figure*}

\begin{figure*}[!t]
\centering
\subfloat[Mip-NeRF360~\cite{barron2022mip}]{\includegraphics[width=2.2in]{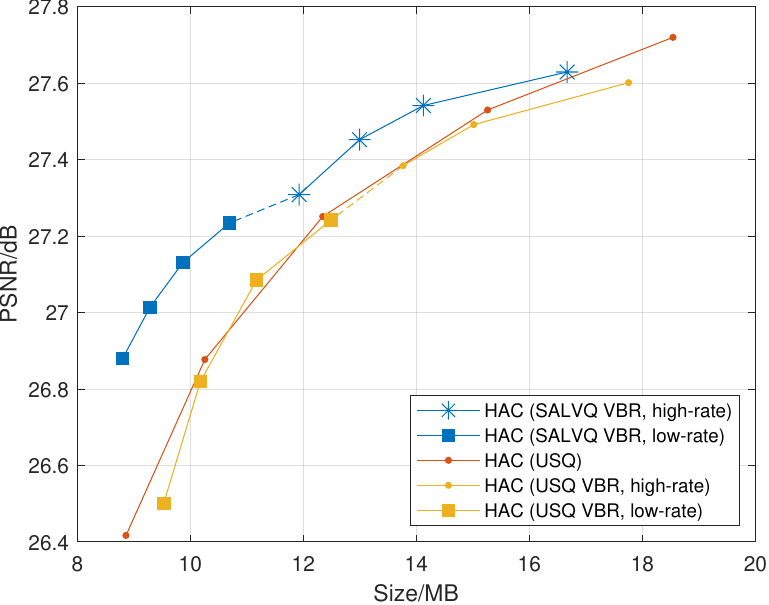}%
}
\hfil
\subfloat[Tank\&Temples~\cite{knapitsch2017tanks}]{\includegraphics[width=2.2in]{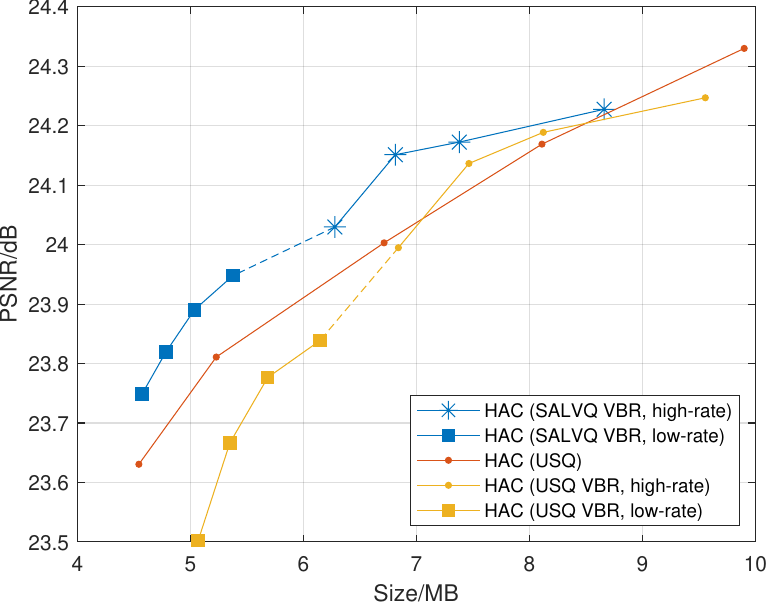}%
}
\hfil
\subfloat[DeepBlending~\cite{hedman2018deep}]{\includegraphics[width=2.2in]{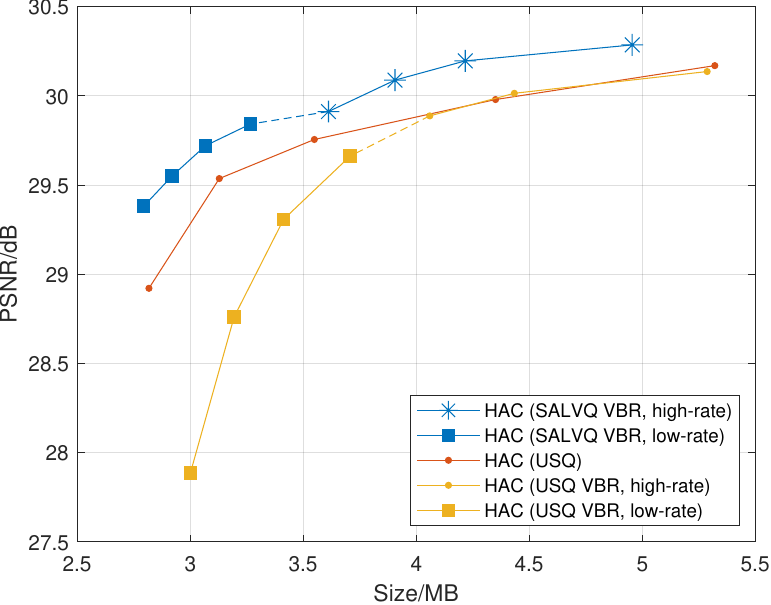}%
}
\caption{Extending the operating bitrate range of variable-rate compression by combining two specialized VBR models for the high-rate and low-rate regions, respectively, using HAC as an example.}
\label{fig:vbr_wider_range}
\end{figure*}

\begin{figure*}[!t]
\centering
\subfloat[Mip-NeRF360~\cite{barron2022mip}]{\includegraphics[width=2.2in]{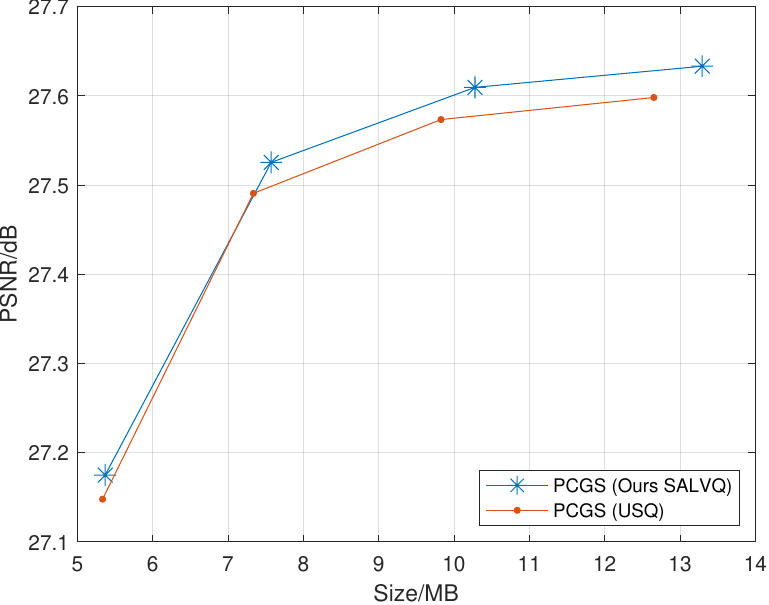}%
}
\hfil
\subfloat[Tank\&Temples~\cite{knapitsch2017tanks}]{\includegraphics[width=2.2in]{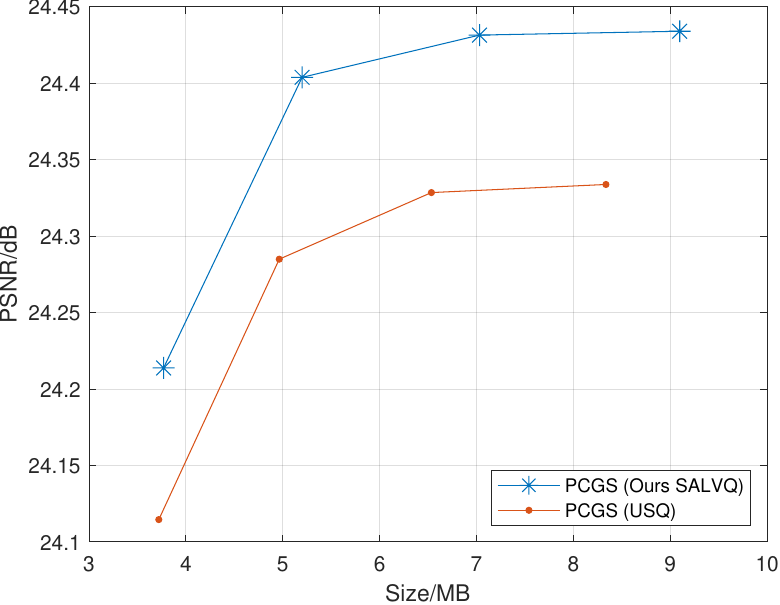}%
}
\hfil
\subfloat[DeepBlending~\cite{hedman2018deep}]{\includegraphics[width=2.2in]{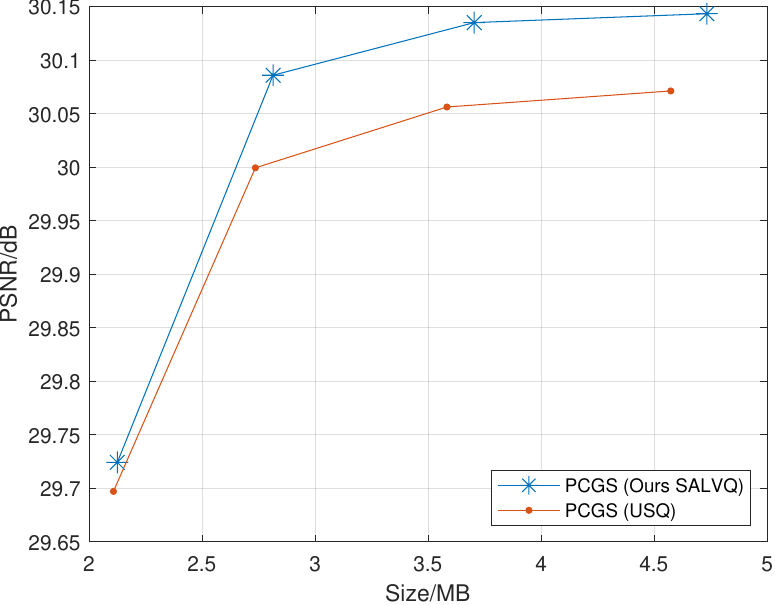}%
}
\caption{Comparison of two quantizers in the PCGS~\cite{chen2025pcgs} architecture}
\label{fig:pcgs_RD_curves}
\end{figure*}

\subsection{Comparison with other 3DGS compression methods}
After completing comparisons among anchor-based compression methods, we further present a comprehensive evaluation against other existing methods, including commonly used baselines~\cite{kerbl20233d,lu2024scaffold,lee2024compact,niedermayr2024compressed,girish2024eagles,fan2023lightgaussian,morgenstern2024compact,navaneet2024compgs,papantonakis2024reducing,wang2024end,liu2024compgs} and the recent feed-forward compression method FCGS~\cite{chen2025fast}. Since these methods typically operate at only one or two bitrate targets, we cannot directly compare with them using R-D curves and BD-rate calculations. Therefore, we use our model trained with $\lambda=0.004$ for evaluation, and report the results in Tab.~\ref{tab:main_results}. For HAC, ContextGS, and HAC++, we additionally include the corresponding USQ-based results in the same table so that readers can directly see the absolute change introduced by replacing USQ with SALVQ at the operating point corresponding to $\lambda=0.004$. We note, however, that for these anchor-based architectures, the more comprehensive comparison between USQ and SALVQ is given by the R--D curves and BD-rate analysis in Fig.~\ref{fig:RD_curves} and Tab.~\ref{tab:lvq_bd_rate}, since they summarize the gain over multiple bitrate targets rather than at a single operating point.

As shown in Tab.~\ref{tab:main_results}, combining our SALVQ approach with any anchor-based compression architecture~\cite{chen2024hac,chen2025hac++,wang2024contextgs} consistently outperforms these baselines in compression performance. On commonly used datasets, including Mip-NeRF 360~\cite{barron2022mip}, Tanks and Temples~\cite{knapitsch2017tanks}, and DeepBlending~\cite{hedman2018deep}, integrating our SALVQ approach with the current SOTA HAC++~\cite{chen2025hac++} achieves an average compression ratio of $131\times$ over the original 3DGS models and $23.7\times$ over Scaffold-GS, while delivering higher or comparable rendering quality. These results further suggest that SALVQ can serve as a generally useful quantization module for anchor-based 3DGS compression architectures.

\subsection{Visual comparison}
We provide visual comparisons between our SALVQ approach and USQ on the ContextGS baseline~\cite{wang2024contextgs} in Fig.~\ref{fig:visual}. The only difference between the two models is the quantizer used, and both are trained with a Lagrange multiplier of $\lambda=0.008$. We report the bitstream size (in MB) and the PSNR (in dB) of the corresponding patches for each case. To demonstrate the effectiveness of the proposed LVQ in reducing rendering distortion, we evaluate four representative scenes: `playroom' (DeepBlending~\cite{hedman2018deep}), `flower' and `stump' (Mip-NeRF360~\cite{barron2022mip}), and `train' (Tanks and Temples~\cite{knapitsch2017tanks}), shown from the first to the fourth row in Fig.~\ref{fig:visual}. In the `playroom' scene, our method better preserves the structure of the wall-mounted switch, which appears appears as an unrecognizable blur in the USQ result. In the `flower' scene, it recovers small flower details that are missing with USQ. For the `stump' scene, our method effectively avoids floater artifacts. In the `train' scene, it better preserves the correct brightness, whereas USQ results are overly bright. The difference in brightness preservation is clearly visible in the residual maps of the Y-channel provided in the last row of Fig.~\ref{fig:visual}.

\begin{table*}
    \centering
    \footnotesize
    \caption{BD-rate of SALVQ-based and USQ-based variable-rate systems relative to that architecture's USQ-based single-rate baseline. \textbf{Bold} numbers highlight cases where variable-rate compression models outperform the corresponding USQ-based single-rate counterparts in R-D performance.}
    %\setlength{\tabcolsep}{3pt}
    %\begin{adjustbox}{width=\linewidth}
    \begin{tabular}{lcccccc}
        \toprule
         &\multicolumn{2}{c}{HAC~\cite{chen2024hac}}&\multicolumn{2}{c}{ContextGS~\cite{wang2024contextgs}}&\multicolumn{2}{c}{HAC++~\cite{chen2025hac++}}\\
         \cline{2-3}\cline{4-5}\cline{6-7}
         &USQ-VBR&SALVQ-VBR&USQ-VBR&SALVQ-VBR&USQ-VBR&SALVQ-VBR\\
         \midrule
         Mip-NeRF360~\cite{barron2022mip}&4.54\% &\textbf{-6.44\%}&2.49\%&\textbf{-5.01\%}&14.99\%&4.16\% \\
         Tank\&Temples~\cite{knapitsch2017tanks}&\textbf{-0.99\%}&\textbf{-13.83\%}&9.87\%&\textbf{-4.65\%}&18.89\%&4.62\%\\
         DeepBlending~\cite{hedman2018deep}&1.74\%&\textbf{-16.89}\%&1.94\%&\textbf{-18.04\%}&4.94\%&\textbf{-8.30}\%\\
         \bottomrule
    \end{tabular}
    %\end{adjustbox}
    \label{tab:variable_rate_model_bd_rate}
\end{table*}
\subsection{Variable-rate compression performance}
\paragraph{Single-model VBR setup}
After evaluating single-rate compression with separately trained models for different target bitrates, we turn to variable-bitrate (VBR) compression, where a single model supports multiple operating points. Since our focus here is on the quantizer itself, we first study single-model VBR achieved through quantization control using the rate-control scheme in Sec.~\ref{sec:rate_control}. Building on this scheme, we compare SALVQ-based and USQ-based VBR systems on HAC, HAC++, and ContextGS. In the current single-model setting, a fixed training budget must be shared across all target rates. As the number of supported operating points increases, the effective optimization budget allocated to each rate decreases, which can noticeably degrade the overall R-D performance. We therefore focus first on the high-rate portion of the R--D curves. Specifically, we use $\lambda \in \{0.002, 0.004, 0.008\}$ and additionally include $\lambda = 0.006$ so that both the VBR model and the corresponding single-rate baseline provide four operating points for BD-rate computation.

\paragraph{Single-model VBR results}
As shown in Fig.~\ref{fig:RD_curves_vbr} and Table~\ref{tab:variable_rate_model_bd_rate}, SALVQ consistently outperforms USQ in VBR mode. Table~\ref{tab:variable_rate_model_bd_rate} reports the BD-rate of USQ-VBR and SALVQ-VBR relative to the corresponding USQ single-rate baseline. In most cases, USQ-VBR incurs a performance loss relative to the USQ single-rate baseline, whereas SALVQ-VBR often matches or even surpasses it; when it does not, it still substantially reduces the loss compared with USQ-VBR. Overall, with four target rates, SALVQ-VBR achieves a better trade-off between rate flexibility and coding performance, while reducing the number of trained models, total training time, and stored model parameters by $4\times$ compared with training four separate single-rate models.

\paragraph{Limitation of a single VBR model and two extensions}
The above results also reveal a practical limitation of the current single-model VBR formulation: regardless of whether SALVQ or USQ is used, a single gain-controlled VBR model is not yet sufficient to cover a broad operating range. Nevertheless, within comparable operating ranges, SALVQ consistently provides better R-D performance than USQ. There are two practical directions to address this issue. The first is to use a small number of region-specific VBR models, i.e., a piecewise VBR strategy, so that each model is specialized to a narrower bitrate region while their combination covers a wider overall range. The second is to treat the number of transmitted Gaussian/anchor primitives as an additional rate-control dimension, jointly with quantization granularity, as exemplified by progressive compression methods such as PCGS~\cite{chen2025pcgs}. We next examine these two directions in turn.

\paragraph{Extending the range with two VBR models}
A simple way to broaden the supported bitrate range is to use a small number of region-specific VBR models. We use HAC as an example in Fig.~\ref{fig:vbr_wider_range}. The USQ-based single-rate baseline is trained at five target rates corresponding to $\lambda \in \{0.002, 0.004, 0.008, 0.015, 0.025\}$. For the piecewise VBR strategy, we train one high-rate VBR model using $\lambda \in \{0.002, 0.004, 0.006, 0.008\}$ and one low-rate VBR model using $\lambda \in \{0.008, 0.012, 0.016, 0.02\}$. Each model supports multiple operating points within its own bitrate region, and their combination substantially broadens the overall bitrate coverage. Compared with training five separate single-rate models, this piecewise VBR strategy reduces the number of trained models to two, corresponding to a $2.5\times$ reduction in training cost. Since each VBR model is trained over a different bitrate region, its R-D curve typically becomes flatter as the bitrate increases within that region. As a result, when two independently trained VBR models are combined, a mild slope discontinuity may appear near the junction of their operating ranges. If the two R-D curves overlap near the junction of their operating ranges, we compare the overlapping operating points from the two models and retain only the one with higher PSNR at a similar bitrate when constructing the final combined R-D curve.

\paragraph{Progressive variable-rate compression}
Another way to extend the operating range is to introduce the number of transmitted Gaussian/anchor primitives as an additional rate-control dimension, jointly with quantization granularity. PCGS is a representative method in this category. To examine whether the gain of SALVQ also carries over to this setting, we replace the USQ module in PCGS with SALVQ and evaluate the resulting system. As shown in Fig.~\ref{fig:pcgs_RD_curves}, SALVQ again yields consistent gains. The BD-rate gains of SALVQ-based PCGS over USQ-based PCGS on Mip-NeRF360, Tanks\&Temples, and DeepBlending are $-3.47\%$, $-20.57\%$, and $-7.01\%$, respectively.

\paragraph{Overall conclusion}
Overall, when a variable-rate 3DGS compression model is required, SALVQ is a better quantizer than USQ, whether the desired bitrate variation lies within a narrow operating range or over an extended operating range. This conclusion holds whether rate control is achieved solely through quantization granularity, through a piecewise VBR strategy with multiple region-specific models, or through the joint use of primitive transmission and quantization control as in PCGS. These results show that SALVQ is effective not only in the standard single-rate setting, but also as a flexible and consistently stronger quantization module for variable-rate 3DGS compression.

\section{Conclusion}
This study investigates the use of LVQ in 3DGS compression. Specifically, we propose a novel SALVQ method, which learns a scene-adaptive lattice basis, making the quantizer's Voronoi cell geometry learnable and aligned to scene statistics. 
SALVQ can be seamlessly integrated into existing 3DGS compression pipelines to improve R-D performance without requiring any modification to other components. It also maintains a computational cost comparable to that of USQ. Furthermore, by adaptively scaling the lattice basis vectors, SALVQ provides effective variable-rate control, often matching or surpassing the R–D performance of USQ-based single-rate models, while eliminating the need to train separate models for different R–D targets. This significantly reduces both training time and memory footprint. Extensive results across multiple architectures show that replacing the USQ module with SALVQ yields consistent gains, positioning SALVQ as a modular, broadly applicable building block for 3DGS compression.

%\section*{Acknowledgments}
%This should be a simple paragraph before the References to thank those individuals and institutions who have supported your work on this article.

%\bibliography{IEEEabrv,../bib/paper}
%

\bibliographystyle{IEEEtran}
\bibliography{main}

@String(CVPR= {IEEE Conf. Comput. Vis. Pattern Recog.})

@String(TOG= {ACM Trans. Graph.})

@String(ICASSP=	{ICASSP})

@String(ICIP = {IEEE Int. Conf. Image Process.})

@String(CVPR  = {CVPR})

@String(TOG   = {ACM TOG})

@String(ICIP  = {ICIP})

@phdthesis{xu2025thesis,
  author       = {Hao Xu},
  title        = {Prior-guided Neural Compression of Visual Data},
  school       = {McMaster University},
  address      = {Hamilton, Canada},
  year         = {2025},
  type         = {{Ph.D.} thesis},
    url          = {http://hdl.handle.net/11375/32258}
}

@book{gersho2012vector,
  title={Vector quantization and signal compression},
  author={Gersho, Allen and Gray, Robert M},
  volume={159},
  year={2012},
  publisher={Springer Science \& Business Media}
}

@inproceedings{xu2024fast,
  title={Fast Point Cloud Geometry Compression with Context-based Residual Coding and INR-based Refinement},
  author={Xu, Hao and Zhang, Xi and Wu, Xiaolin},
  booktitle={European Conference on Computer Vision},
  pages={270--288},
  year={2024},
  organization={Springer}
}

@inproceedings{he2022density,
  title={Density-preserving deep point cloud compression},
  author={He, Yun and Ren, Xinlin and Tang, Danhang and Zhang, Yinda and Xue, Xiangyang and Fu, Yanwei},
  booktitle={Proceedings of the IEEE/CVF Conference on Computer Vision and Pattern Recognition},
  pages={2333--2342},
  year={2022}
}

@article{de2016compression,
  title={Compression of 3D point clouds using a region-adaptive hierarchical transform},
  author={De Queiroz, Ricardo L and Chou, Philip A},
  journal={IEEE Transactions on Image Processing},
  volume={25},
  number={8},
  pages={3947--3956},
  year={2016},
  publisher={IEEE}
}

@inproceedings{zhang2014point,
  title={Point cloud attribute compression with graph transform},
  author={Zhang, Cha and Florencio, Dinei and Loop, Charles},
  booktitle={2014 IEEE International Conference on Image Processing (ICIP)},
  pages={2066--2070},
  year={2014},
  organization={IEEE}
}

@inproceedings{wang2022sparse,
  title={Sparse tensor-based point cloud attribute compression},
  author={Wang, Jianqiang and Ma, Zhan},
  booktitle={2022 IEEE 5th International Conference on Multimedia Information Processing and Retrieval (MIPR)},
  pages={59--64},
  year={2022},
  organization={IEEE}
}

@article{gu20193d,
  title={3D point cloud attribute compression using geometry-guided sparse representation},
  author={Gu, Shuai and Hou, Junhui and Zeng, Huanqiang and Yuan, Hui and Ma, Kai-Kuang},
  journal={IEEE Transactions on Image Processing},
  volume={29},
  pages={796--808},
  year={2019},
  publisher={IEEE}
}

@article{chou2019volumetric,
  title={A volumetric approach to point cloud compression—Part I: Attribute compression},
  author={Chou, Philip A and Koroteev, Maxim and Krivoku{\'c}a, Maja},
  journal={IEEE Transactions on Image Processing},
  volume={29},
  pages={2203--2216},
  year={2019},
  publisher={IEEE}
}

@article{sheng2022attribute,
  title={Attribute artifacts removal for geometry-based point cloud compression},
  author={Sheng, Xihua and Li, Li and Liu, Dong and Xiong, Zhiwei},
  journal={IEEE Transactions on Image Processing},
  volume={31},
  pages={3399--3413},
  year={2022},
  publisher={IEEE}
}

@article{lombardi2019neural,
	title={Neural volumes: Learning dynamic renderable volumes from images},
	author={Lombardi, Stephen and Simon, Tomas and Saragih, Jason and Schwartz, Gabriel and Lehrmann, Andreas and Sheikh, Yaser},
	journal={arXiv preprint arXiv:1906.07751},
	year={2019}
}

@article{mildenhall2019local,
	title={Local light field fusion: Practical view synthesis with prescriptive sampling guidelines},
	author={Mildenhall, Ben and Srinivasan, Pratul P and Ortiz-Cayon, Rodrigo and Kalantari, Nima Khademi and Ramamoorthi, Ravi and Ng, Ren and Kar, Abhishek},
	journal={ACM Transactions on Graphics (ToG)},
	volume={38},
	number={4},
	pages={1--14},
	year={2019},
	publisher={ACM New York, NY, USA}
}

@article{sitzmann2019scene,
	title={Scene representation networks: Continuous 3d-structure-aware neural scene representations},
	author={Sitzmann, Vincent and Zollh{\"o}fer, Michael and Wetzstein, Gordon},
	journal={Advances in neural information processing systems},
	volume={32},
	year={2019}
}

@article{mildenhall2021nerf,
	title={Nerf: Representing scenes as neural radiance fields for view synthesis},
	author={Mildenhall, Ben and Srinivasan, Pratul P and Tancik, Matthew and Barron, Jonathan T and Ramamoorthi, Ravi and Ng, Ren},
	journal={Communications of the ACM},
	volume={65},
	number={1},
	pages={99--106},
	year={2021},
	publisher={ACM New York, NY, USA}
}

@article{kerbl20233d,
	title={3d gaussian splatting for real-time radiance field rendering.},
	author={Kerbl, Bernhard and Kopanas, Georgios and Leimk{\"u}hler, Thomas and Drettakis, George},
	journal={ACM Trans. Graph.},
	volume={42},
	number={4},
	pages={139--1},
	year={2023}
}

@inproceedings{lu2024scaffold,
	title={Scaffold-gs: Structured 3d gaussians for view-adaptive rendering},
	author={Lu, Tao and Yu, Mulin and Xu, Linning and Xiangli, Yuanbo and Wang, Limin and Lin, Dahua and Dai, Bo},
	booktitle={Proceedings of the IEEE/CVF Conference on Computer Vision and Pattern Recognition},
	pages={20654--20664},
	year={2024}
}

@inproceedings{morgenstern2024compact,
	title={Compact 3d scene representation via self-organizing gaussian grids},
	author={Morgenstern, Wieland and Barthel, Florian and Hilsmann, Anna and Eisert, Peter},
	booktitle={European Conference on Computer Vision},
	pages={18--34},
	year={2024},
	organization={Springer}
}

@article{liu2024hemgs,
	title={HEMGS: A Hybrid Entropy Model for 3D Gaussian Splatting Data Compression},
	author={Liu, Lei and Chen, Zhenghao and Xu, Dong},
	journal={arXiv preprint arXiv:2411.18473},
	year={2024}
}

@inproceedings{
	zhan2025catdgs,
	title={{CAT}-3{DGS}: A Context-Adaptive Triplane Approach to Rate-Distortion-Optimized 3{DGS} Compression},
	author={Yu-Ting Zhan and Cheng-Yuan Ho and Hebi Yang and Yi-Hsin Chen and Jui Chiu Chiang and Yu-Lun Liu and Wen-Hsiao Peng},
	booktitle={The Thirteenth International Conference on Learning Representations},
	year={2025},
	url={https://openreview.net/forum?id=m3KuuE2ozw}
}

@inproceedings{chen2024hac,
	title={Hac: Hash-grid assisted context for 3d gaussian splatting compression},
	author={Chen, Yihang and Wu, Qianyi and Lin, Weiyao and Harandi, Mehrtash and Cai, Jianfei},
	booktitle={European Conference on Computer Vision},
	pages={422--438},
	year={2024},
	organization={Springer}
}

@inproceedings{liu2024compgs,
	title={Compgs: Efficient 3d scene representation via compressed gaussian splatting},
	author={Liu, Xiangrui and Wu, Xinju and Zhang, Pingping and Wang, Shiqi and Li, Zhu and Kwong, Sam},
	booktitle={Proceedings of the 32nd ACM International Conference on Multimedia},
	pages={2936--2944},
	year={2024}
}

@article{wang2024contextgs,
  title={Contextgs: Compact 3d gaussian splatting with anchor level context model},
  author={Wang, Yufei and Li, Zhihao and Guo, Lanqing and Yang, Wenhan and Kot, Alex and Wen, Bihan},
  journal={Advances in neural information processing systems},
  volume={37},
  pages={51532--51551},
  year={2024}
}

@inproceedings{girish2024eagles,
	title={Eagles: Efficient accelerated 3d gaussians with lightweight encodings},
	author={Girish, Sharath and Gupta, Kamal and Shrivastava, Abhinav},
	booktitle={European Conference on Computer Vision},
	pages={54--71},
	year={2024},
	organization={Springer}
}

@inproceedings{xie2024mesongs,
	title={Mesongs: Post-training compression of 3d gaussians via efficient attribute transformation},
	author={Xie, Shuzhao and Zhang, Weixiang and Tang, Chen and Bai, Yunpeng and Lu, Rongwei and Ge, Shijia and Wang, Zhi},
	booktitle={European Conference on Computer Vision},
	pages={434--452},
	year={2024},
	organization={Springer}
}

@article{papantonakis2024reducing,
	title={Reducing the Memory Footprint of 3D Gaussian Splatting},
	author={Papantonakis, Panagiotis and Kopanas, Georgios and Kerbl, Bernhard and Lanvin, Alexandre and Drettakis, George},
	journal={Proceedings of the ACM on Computer Graphics and Interactive Techniques},
	volume={7},
	number={1},
	pages={1--17},
	year={2024},
	publisher={ACM New York, NY, USA}
}

@inproceedings{wang2024end,
	title={End-to-end rate-distortion optimized 3d gaussian representation},
	author={Wang, Henan and Zhu, Hanxin and He, Tianyu and Feng, Runsen and Deng, Jiajun and Bian, Jiang and Chen, Zhibo},
	booktitle={European Conference on Computer Vision},
	pages={76--92},
	year={2024},
	organization={Springer}
}

@article{fan2023lightgaussian,
	title={Lightgaussian: Unbounded 3d gaussian compression with 15x reduction and 200+ fps},
	author={Fan, Zhiwen and Wang, Kevin and Wen, Kairun and Zhu, Zehao and Xu, Dejia and Wang, Zhangyang},
	journal={arXiv preprint arXiv:2311.17245},
	year={2023}
}

@inproceedings{lee2024compact,
	title={Compact 3d gaussian representation for radiance field},
	author={Lee, Joo Chan and Rho, Daniel and Sun, Xiangyu and Ko, Jong Hwan and Park, Eunbyung},
	booktitle={Proceedings of the IEEE/CVF Conference on Computer Vision and Pattern Recognition},
	pages={21719--21728},
	year={2024}
}

@inproceedings{niedermayr2024compressed,
	title={Compressed 3d gaussian splatting for accelerated novel view synthesis},
	author={Niedermayr, Simon and Stumpfegger, Josef and Westermann, R{\"u}diger},
	booktitle={Proceedings of the IEEE/CVF Conference on Computer Vision and Pattern Recognition},
	pages={10349--10358},
	year={2024}
}

@inproceedings{navaneet2024compgs,
	title={Compgs: Smaller and faster gaussian splatting with vector quantization},
	author={Navaneet, KL and Pourahmadi Meibodi, Kossar and Abbasi Koohpayegani, Soroush and Pirsiavash, Hamed},
	booktitle={European Conference on Computer Vision},
	pages={330--349},
	year={2024},
	organization={Springer}
}

@article{niemeyer2024radsplat,
	title={Radsplat: Radiance field-informed gaussian splatting for robust real-time rendering with 900+ fps},
	author={Niemeyer, Michael and Manhardt, Fabian and Rakotosaona, Marie-Julie and Oechsle, Michael and Duckworth, Daniel and Gosula, Rama and Tateno, Keisuke and Bates, John and Kaeser, Dominik and Tombari, Federico},
	journal={arXiv preprint arXiv:2403.13806},
	year={2024}
}

@article{ali2024elmgs,
	title={ELMGS: Enhancing memory and computation scaLability through coMpression for 3D Gaussian Splatting},
	author={Ali, Muhammad Salman and Bae, Sung-Ho and Tartaglione, Enzo},
	journal={arXiv preprint arXiv:2410.23213},
	year={2024}
}

@article{hanson2024pup,
	title={PUP 3D-GS: Principled Uncertainty Pruning for 3D Gaussian Splatting},
	author={Hanson, Alex and Tu, Allen and Singla, Vasu and Jayawardhana, Mayuka and Zwicker, Matthias and Goldstein, Tom},
	journal={arXiv preprint arXiv:2406.10219},
	year={2024}
}

@article{liu2024efficientgs,
	title={EfficientGS: Streamlining Gaussian Splatting for Large-Scale High-Resolution Scene Representation},
	author={Liu, Wenkai and Guan, Tao and Zhu, Bin and Ju, Lili and Song, Zikai and Li, Dan and Wang, Yuesong and Yang, Wei},
	journal={arXiv preprint arXiv:2404.12777},
	year={2024}
}

@article{lee2024safeguardgs,
	title={SafeguardGS: 3D Gaussian Primitive Pruning While Avoiding Catastrophic Scene Destruction},
	author={Lee, Yongjae and Zhang, Zhaoliang and Fan, Deliang},
	journal={arXiv preprint arXiv:2405.17793},
	year={2024}
}

@inproceedings{fang2024mini,
	title={Mini-splatting: Representing scenes with a constrained number of gaussians},
	author={Fang, Guangchi and Wang, Bing},
	booktitle={European Conference on Computer Vision},
	pages={165--181},
	year={2024},
	organization={Springer}
}

@book{sayood2017introduction,
	title={Introduction to data compression},
	author={Sayood, Khalid},
	year={2017},
	publisher={Morgan Kaufmann}
}

@book{conway2013sphere,
	title={Sphere packings, lattices and groups},
	author={Conway, John Horton and Sloane, Neil James Alexander},
	volume={290},
	year={2013},
	publisher={Springer Science \& Business Media}
}

@article{conway1982fast,
	title={Fast quantizing and decoding and algorithms for lattice quantizers and codes},
	author={Conway, John and Sloane, Neil},
	journal={IEEE Transactions on Information Theory},
	volume={28},
	number={2},
	pages={227--232},
	year={1982},
	publisher={IEEE}
}

@ARTICLE{chen2025hac++,
  author={Chen, Yihang and Wu, Qianyi and Lin, Weiyao and Harandi, Mehrtash and Cai, Jianfei},
  journal={IEEE Transactions on Pattern Analysis and Machine Intelligence}, 
  title={HAC++: Towards 100X Compression of 3D Gaussian Splatting}, 
  year={2025},
  volume={},
  number={},
  pages={1-17},
  doi={10.1109/TPAMI.2025.3594066}}

@article{van2017neural,
	title={Neural discrete representation learning},
	author={Van Den Oord, Aaron and Vinyals, Oriol and others},
	journal={Advances in neural information processing systems},
	volume={30},
	year={2017}
}

@article{agustsson2017soft,
	title={Soft-to-hard vector quantization for end-to-end learning compressible representations},
	author={Agustsson, Eirikur and Mentzer, Fabian and Tschannen, Michael and Cavigelli, Lukas and Timofte, Radu and Benini, Luca and Gool, Luc V},
	journal={Advances in neural information processing systems},
	volume={30},
	year={2017}
}

@article{bengio2013estimating,
	title={Estimating or propagating gradients through stochastic neurons for conditional computation},
	author={Bengio, Yoshua and L{\'e}onard, Nicholas and Courville, Aaron},
	journal={arXiv preprint arXiv:1308.3432},
	year={2013}
}

@article{takida2022sq,
	title={Sq-vae: Variational bayes on discrete representation with self-annealed stochastic quantization},
	author={Takida, Yuhta and Shibuya, Takashi and Liao, WeiHsiang and Lai, Chieh-Hsin and Ohmura, Junki and Uesaka, Toshimitsu and Murata, Naoki and Takahashi, Shusuke and Kumakura, Toshiyuki and Mitsufuji, Yuki},
	journal={arXiv preprint arXiv:2205.07547},
	year={2022}
}

@inproceedings{
chen2025fast,
title={Fast Feedforward 3D Gaussian Splatting Compression},
author={Yihang Chen and Qianyi Wu and Mengyao Li and Weiyao Lin and Mehrtash Harandi and Jianfei Cai},
booktitle={The Thirteenth International Conference on Learning Representations},
year={2025},
url={https://openreview.net/forum?id=DCandSZ2F1}
}

@article{knapitsch2017tanks,
	title={Tanks and temples: Benchmarking large-scale scene reconstruction},
	author={Knapitsch, Arno and Park, Jaesik and Zhou, Qian-Yi and Koltun, Vladlen},
	journal={ACM Transactions on Graphics (ToG)},
	volume={36},
	number={4},
	pages={1--13},
	year={2017},
	publisher={ACM New York, NY, USA}
}

@article{hedman2018deep,
	title={Deep blending for free-viewpoint image-based rendering},
	author={Hedman, Peter and Philip, Julien and Price, True and Frahm, Jan-Michael and Drettakis, George and Brostow, Gabriel},
	journal={ACM Transactions on Graphics (ToG)},
	volume={37},
	number={6},
	pages={1--15},
	year={2018},
	publisher={ACM New York, NY, USA}
}

@inproceedings{barron2022mip,
	title={Mip-nerf 360: Unbounded anti-aliased neural radiance fields},
	author={Barron, Jonathan T and Mildenhall, Ben and Verbin, Dor and Srinivasan, Pratul P and Hedman, Peter},
	booktitle={Proceedings of the IEEE/CVF conference on computer vision and pattern recognition},
	pages={5470--5479},
	year={2022}
}

@article{minnen2018joint,
  title={Joint autoregressive and hierarchical priors for learned image compression},
  author={Minnen, David and Ball{\'e}, Johannes and Toderici, George D},
  journal={Advances in neural information processing systems},
  volume={31},
  year={2018}
}

@inproceedings{minnen2020channel,
	title={Channel-wise autoregressive entropy models for learned image compression},
	author={Minnen, David and Singh, Saurabh},
	booktitle={2020 IEEE International Conference on Image Processing (ICIP)},
	pages={3339--3343},
	year={2020},
	organization={IEEE}
}

@inproceedings{he2021checkerboard,
	title={Checkerboard context model for efficient learned image compression},
	author={He, Dailan and Zheng, Yaoyan and Sun, Baocheng and Wang, Yan and Qin, Hongwei},
	booktitle={Proceedings of the IEEE/CVF Conference on Computer Vision and Pattern Recognition},
	pages={14771--14780},
	year={2021}
}

@inproceedings{zhang2023lvqac,
	title={Lvqac: Lattice vector quantization coupled with spatially adaptive companding for efficient learned image compression},
	author={Zhang, Xi and Wu, Xiaolin},
	booktitle={Proceedings of the IEEE/CVF Conference on Computer Vision and Pattern Recognition},
	pages={10239--10248},
	year={2023}
}

@inproceedings{
lei2025approaching,
title={Approaching Rate-Distortion Limits in Neural Compression with Lattice Transform Coding},
author={Eric Lei and Hamed Hassani and Shirin Saeedi Bidokhti},
booktitle={The Thirteenth International Conference on Learning Representations},
year={2025},
url={https://openreview.net/forum?id=Tv36j85SqR}
}

@article{bjontegaard2001calculation,
	title={Calculation of average PSNR differences between RD-curves},
	author={Bjontegaard, Gisle},
	journal={ITU SG16 Doc. VCEG-M33},
	year={2001}
}

@article{wang2004image,
	title={Image quality assessment: from error visibility to structural similarity},
	author={Wang, Zhou and Bovik, Alan C and Sheikh, Hamid R and Simoncelli, Eero P},
	journal={IEEE transactions on image processing},
	volume={13},
	number={4},
	pages={600--612},
	year={2004},
	publisher={IEEE}
}

@inproceedings{zhang2018unreasonable,
	title={The unreasonable effectiveness of deep features as a perceptual metric},
	author={Zhang, Richard and Isola, Phillip and Efros, Alexei A and Shechtman, Eli and Wang, Oliver},
	booktitle={Proceedings of the IEEE conference on computer vision and pattern recognition},
	pages={586--595},
	year={2018}
}

@article{babai1986lovasz,
  title={On Lov{\'a}sz’lattice reduction and the nearest lattice point problem},
  author={Babai, L{\'a}szl{\'o}},
  journal={Combinatorica},
  volume={6},
  pages={1--13},
  year={1986},
  publisher={Springer}
}

@article{zhang2024learning,
  title={Learning optimal lattice vector quantizers for end-to-end neural image compression},
  author={Zhang, Xi and Wu, Xiaolin},
  journal={Advances in Neural Information Processing Systems},
  volume={37},
  pages={106497--106518},
  year={2024}
}

@inproceedings{cao2024entropy,
  title={Entropy Relaxed Lattice Vector Quantization for Learned Image Compression},
  author={Cao, Maida and Dai, Wenrui and Li, Shaohui and Li, Han and Li, Chenglin and Zou, Junni and Xiong, Hongkai},
  booktitle={2024 Data Compression Conference (DCC)},
  pages={548--548},
  year={2024},
  organization={IEEE}
}

@InProceedings{Xu_2025_CVPR,
    author    = {Xu, Hao and Wu, Xiaolin and Zhang, Xi},
    title     = {Multirate Neural Image Compression with Adaptive Lattice Vector Quantization},
    booktitle = {Proceedings of the Computer Vision and Pattern Recognition Conference (CVPR)},
    month     = {June},
    year      = {2025},
    pages     = {7633-7642}
}

@article{paszke2019pytorch,
  title={Pytorch: An imperative style, high-performance deep learning library},
  author={Paszke, Adam and Gross, Sam and Massa, Francisco and Lerer, Adam and Bradbury, James and Chanan, Gregory and Killeen, Trevor and Lin, Zeming and Gimelshein, Natalia and Antiga, Luca and others},
  journal={Advances in neural information processing systems},
  volume={32},
  year={2019}
}

@inproceedings{chen2020variable,
  title={Variable bitrate image compression with quality scaling factors},
  author={Chen, Tong and Ma, Zhan},
  booktitle={ICASSP 2020-2020 IEEE International Conference on Acoustics, Speech and Signal Processing (ICASSP)},
  pages={2163--2167},
  year={2020},
  organization={IEEE}
}

@inproceedings{cui2021asymmetric,
  title={Asymmetric gained deep image compression with continuous rate adaptation},
  author={Cui, Ze and Wang, Jing and Gao, Shangyin and Guo, Tiansheng and Feng, Yihui and Bai, Bo},
  booktitle={Proceedings of the IEEE/CVF Conference on Computer Vision and Pattern Recognition},
  pages={10532--10541},
  year={2021}
}

@inproceedings{kamisli2024variable,
  title={Variable-Rate Learned Image Compression with Multi-Objective Optimization and Quantization-Reconstruction Offsets},
  author={Kamisli, Fatih and Racap{\'e}, Fabien and Choi, Hyomin},
  booktitle={2024 Data Compression Conference (DCC)},
  pages={193--202},
  year={2024},
  organization={IEEE}
}

@inproceedings{tong2023qvrf,
  title={Qvrf: A quantization-error-aware variable rate framework for learned image compression},
  author={Tong, Kedeng and Wu, Yaojun and Li, Yue and Zhang, Kai and Zhang, Li and Jin, Xin},
  booktitle={2023 IEEE International Conference on Image Processing (ICIP)},
  pages={1310--1314},
  year={2023},
  organization={IEEE}
}

@article{chen2025pcgs,
  title={Pcgs: Progressive compression of 3d gaussian splatting},
  author={Chen, Yihang and Li, Mengyao and Wu, Qianyi and Lin, Weiyao and Harandi, Mehrtash and Cai, Jianfei},
  journal={arXiv preprint arXiv:2503.08511},
  year={2025}
}

\begin{IEEEbiography}
  [{\includegraphics[width=1in,clip,keepaspectratio]{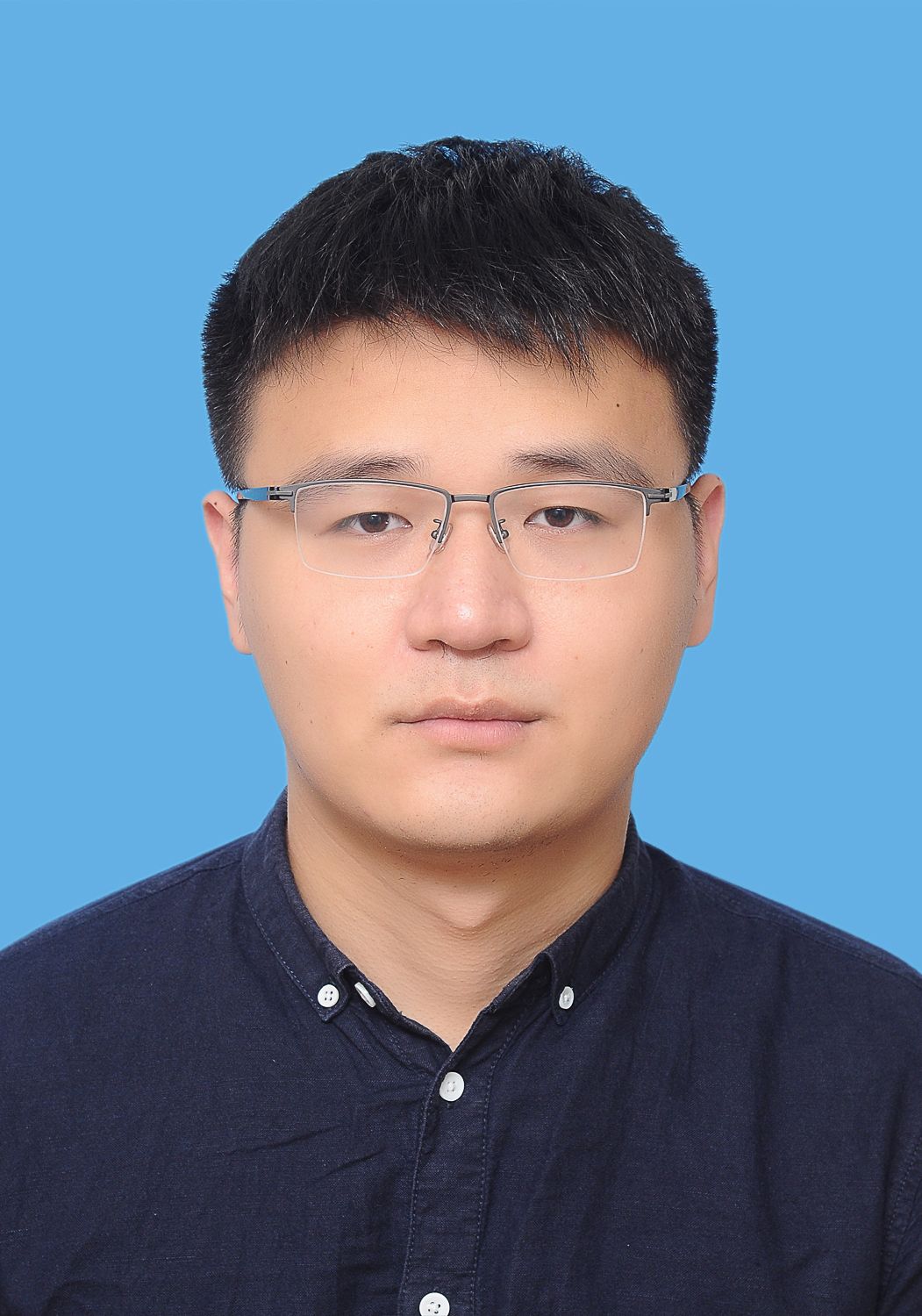}}]{Hao Xu} received the B.Eng. degree in Communication Engineering from Shandong University, Qingdao, China, in 2021, and the Ph.D. degree in Electrical and Computer Engineering from McMaster University, Hamilton, ON, Canada, in August 2025. He is currently a Postdoctoral Fellow at McMaster University, working with Dr. Xiaolin Wu. His research focuses on neural data compression and memory-efficient 3D scene representations, spanning the compression of images, point clouds, and 3D/4D Gaussian Splatting. He is also interested in efficient AI, particularly efficient Vision Transformers and computationally efficient feedforward 3D/4D models.
\end{IEEEbiography}

\begin{IEEEbiography}
  [{\includegraphics[width=1in,clip,keepaspectratio]{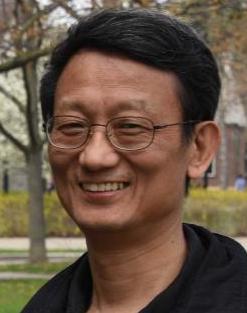}}]{Xiaolin Wu}
  (Life Fellow, IEEE) 
  received the B.Sc. degree in computer science 
  from Wuhan University, China, in 1982, and the Ph.D. degree in computer science 
  from the University of Calgary, Canada, in 1988. He started his academic career 
  in 1988. He was a Faculty Member with Western University, Canada, and New York 
  Polytechnic University (NYU-Poly). He is currently with McMaster University and 
  Southwest Jiaotong University. His research interests include image processing, 
  data compression, digital multimedia, low-level vision, and network-aware visual 
  communication. He has authored or co-authored more than 300 research articles and
  holds four patents in these fields. He served on technical committees for many 
  IEEE international conferences/workshops on image processing, multimedia, data 
  compression, and information theory. He was a past Associate Editor of IEEE 
  Transactions on Multimedia and IEEE Transactions on Image Processing.
\end{IEEEbiography}

\begin{IEEEbiography}
  [{\includegraphics[width=1in,clip,keepaspectratio]{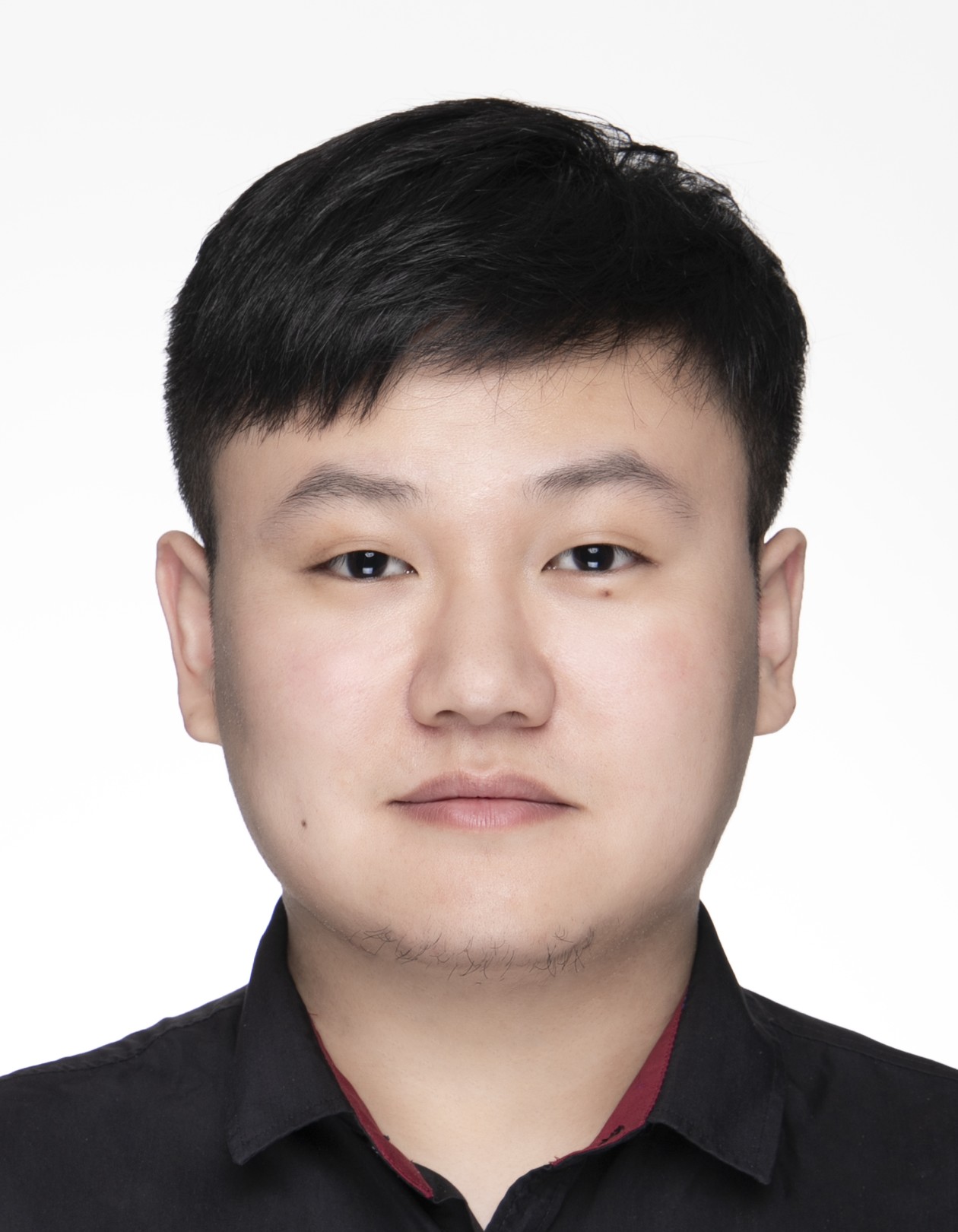}}]{Xi Zhang}
  (Member, IEEE) 
  received the B.Sc. degree in mathematics and physics basic science from the University of Electronic Science and Technology of China in 2015, and the Ph.D. degree in electronic engineering from Shanghai Jiao Tong University, China, in 2022. He was a Postdoctoral Fellow with McMaster University, Canada, from July 2022 to August 2024, and a Research Scientist with Nanyang Technological University, Singapore, from September 2024 to May 2026. He is currently an Associate Professor with the School of Computer Science and Technology, Tongji University, China. His research interests include green and energy-efficient artificial intelligence, learning-based data compression, visual signal processing, and efficient AI systems.
\end{IEEEbiography}

\clearpage

\twocolumn[
\begin{center}
  \textbf{{\Large --Supplementary Material--}}
\end{center}
\vspace{15pt}
]
\begin{table}[h!]\small
    \centering
    \setlength{\abovecaptionskip}{0cm}
	\setlength{\belowcaptionskip}{0.2cm}
    \caption{Results of `HAC~\cite{chen2024hac} + Ours SALVQ' for each scene from Deep Blending dataset~\cite{hedman2018deep}.}
    \begin{adjustbox}{width=\linewidth}
    \begin{tabular}{m{3cm}<{\centering} |c|c c c c}
        % \hline
        \toprule[1.2pt]
        \textbf{\(\lambda\)} & \textbf{Scenes} & \textbf{PSNR}↑ & \textbf{SSIM}↑ & \textbf{LPIPS}↓ & \textbf{SIZE}↓ \\ 
        
        \hline
        \multirow{3}{*}{0.002}
        & drjohnson & 29.72 & 0.905 & 0.258 & 6.15 \\
        & playroom & 30.63 & 0.904 & 0.265 & 3.79 \\
        & \textbf{AVG} & \textbf{30.18} & \textbf{0.905} & \textbf{0.261} & \textbf{4.97} \\
        \hline
        \multirow{3}{*}{0.004}
        & drjohnson & 29.61 & 0.904 & 0.263 & 4.93 \\
        & playroom & 30.51 & 0.902 & 0.273 & 3.03 \\
        & \textbf{AVG} & \textbf{30.06} & \textbf{0.903} & \textbf{0.268} & \textbf{3.98} \\
        \hline
        \multirow{3}{*}{0.008}
        & drjohnson & 29.39 & 0.899 & 0.274 & 3.99 \\
        & playroom & 30.29 & 0.899 & 0.278 & 2.56 \\
        & \textbf{AVG} & \textbf{29.84} & \textbf{0.899} & \textbf{0.276} & \textbf{3.27} \\
        \hline
        \multirow{3}{*}{0.015}
        & drjohnson & 29.04 & 0.898 & 0.279 & 3.45 \\
        & playroom & 30.20 & 0.903 & 0.280 & 2.18 \\
        & \textbf{AVG} & \textbf{29.62} & \textbf{0.901} & \textbf{0.280} & \textbf{2.81} \\
        \hline
        \multirow{3}{*}{0.025}
        & drjohnson & 28.96 & 0.895 & 0.287 & 3.18 \\
        & playroom & 29.73 & 0.889 & 0.290 & 1.94 \\
        & \textbf{AVG} & \textbf{29.35} & \textbf{0.897} & \textbf{0.288} & \textbf{2.56} \\
        \bottomrule[1.2pt]
    \end{tabular}
    \end{adjustbox}
\label{table: our_db_hac}
\end{table}

\begin{table}[!ht]\small
    \centering
    \setlength{\abovecaptionskip}{0cm}
	\setlength{\belowcaptionskip}{0.2cm}
    \caption{Results of `HAC~\cite{chen2024hac} + Ours SALVQ' for each scene from Tank \& Temples dataset~\cite{knapitsch2017tanks}.}
    \begin{adjustbox}{width=\linewidth}
    \begin{tabular}{m{3cm}<{\centering} |c|c c c c}
        % \hline
        \toprule[1.2pt]
        \textbf{\(\lambda_r\)} & \textbf{Scenes} & \textbf{PSNR}↑ & \textbf{SSIM}↑ & \textbf{LPIPS}↓ & \textbf{SIZE}↓ \\ 
        
        \hline
        \multirow{3}{*}{0.002}
        & train & 22.69 & 0.820 & 0.210 & 8.10 \\
        & truck & 25.99 & 0.881 & 0.150 & 10.06 \\
        & \textbf{AVG} & \textbf{24.34} & \textbf{0.850} & \textbf{0.180} & \textbf{9.08} \\
        \hline
        \multirow{3}{*}{0.004}
        & train & 22.47 & 0.816 & 0.215 & 6.52 \\
        & truck & 25.92 & 0.878 & 0.157 & 8.02 \\
        & \textbf{AVG} & \textbf{24.20} & \textbf{0.847} & \textbf{0.186} & \textbf{7.27} \\
        \hline
        \multirow{3}{*}{0.008}
        & train & 22.42 & 0.809 & 0.227 & 4.82 \\
        & truck & 25.75 & 0.873 & 0.167 & 6.92 \\
        & \textbf{AVG} & \textbf{24.09} & \textbf{0.841} & \textbf{0.197} & \textbf{5.92} \\
        \hline
        \multirow{3}{*}{0.015}
        & train & 22.22 & 0.802 & 0.238 & 4.02 \\
        & truck & 25.58 & 0.871 & 0.176 & 5.66 \\
        & \textbf{AVG} & \textbf{23.90} & \textbf{0.837} & \textbf{0.207} & \textbf{4.84} \\
        \hline
        \multirow{3}{*}{0.025}
        & train & 22.06 & 0.792 & 0.253 & 3.19 \\
        & truck & 25.33 & 0.863 & 0.191 & 4.99 \\
        & \textbf{AVG} & \textbf{23.70} & \textbf{0.827} & \textbf{0.222} & \textbf{4.09} \\
       
        \bottomrule[1.2pt]
    \end{tabular}
    \end{adjustbox}
\label{table: our_tnt_hac}
\end{table}

\begin{table}[!ht]\small
    \centering
    \setlength{\abovecaptionskip}{0cm}
	\setlength{\belowcaptionskip}{0.2cm}
    \caption{Results of `HAC~\cite{chen2024hac} + Ours SALVQ' for each scene from Mip-NeRF 360 dataset~\cite{barron2022mip}.}
    \begin{adjustbox}{width=\linewidth}
    \begin{tabular}{m{3cm}<{\centering} |c|c c c c}
        % \hline
        \toprule[1.2pt]
        \textbf{\(\lambda\)} & \textbf{Scenes} & \textbf{PSNR}↑ & \textbf{SSIM}↑ & \textbf{LPIPS}↓ & \textbf{SIZE}↓ \\ 
        \hline
        \multirow{10}{*}{0.002}
        & bicycle  & 25.15 & 0.744 & 0.261 & 32.93 \\
        & bonsai   & 32.83 & 0.946 & 0.183 &  9.74 \\
        & counter  & 29.62 & 0.915 & 0.188 &  8.24 \\
        & flowers  & 21.31 & 0.575 & 0.378 & 22.19 \\
        & garden   & 27.42 & 0.847 & 0.143 & 26.16 \\
        & kitchen  & 31.57 & 0.928 & 0.124 &  9.50 \\
        & room     & 32.02 & 0.925 & 0.200 &  5.86 \\
        & stump    & 26.65 & 0.763 & 0.265 & 20.57 \\
        & treehill & 23.36 & 0.646 & 0.351 & 23.97 \\
        & \textbf{AVG} & \textbf{27.77} & \textbf{0.810} & \textbf{0.233} & \textbf{17.68} \\
        \hline
        \multirow{10}{*}{0.004}
        & bicycle  & 25.06 & 0.742 & 0.267 & 26.80 \\
        & bonsai   & 32.40 & 0.943 & 0.188 &  7.94 \\
        & counter  & 29.47 & 0.911 & 0.195 &  6.71 \\
        & flowers  & 21.29 & 0.574 & 0.380 & 18.46 \\
        & garden   & 27.31 & 0.842 & 0.153 & 21.12 \\
        & kitchen  & 31.27 & 0.925 & 0.130 &  7.64 \\
        & room     & 31.73 & 0.922 & 0.208 &  4.79 \\
        & stump    & 26.60 & 0.760 & 0.273 & 16.66 \\
        & treehill & 23.29 & 0.644 & 0.358 & 18.62 \\
        & \textbf{AVG} & \textbf{27.60} & \textbf{0.807} & \textbf{0.239} & \textbf{14.30} \\
        \hline
        \multirow{10}{*}{0.008}
        & bicycle  & 24.94 & 0.736 & 0.275 & 21.60 \\
        & bonsai   & 31.85 & 0.937 & 0.197 &  6.63 \\
        & counter  & 29.07 & 0.905 & 0.204 &  5.34 \\
        & flowers  & 21.17 & 0.564 & 0.392 & 14.25 \\
        & garden   & 27.05 & 0.832 & 0.167 & 16.35 \\
        & kitchen  & 30.80 & 0.918 & 0.139 &  5.94 \\
        & room     & 31.48 & 0.917 & 0.217 &  3.95 \\
        & stump    & 26.51 & 0.756 & 0.283 & 13.66 \\
        & treehill & 23.19 & 0.638 & 0.370 & 14.71 \\
        & \textbf{AVG} & \textbf{27.34} & \textbf{0.800} & \textbf{0.249} & \textbf{11.38} \\
        \hline
        \multirow{10}{*}{0.015}
        & bicycle  & 24.70 & 0.727 & 0.289 & 17.45 \\
        & bonsai   & 31.26 & 0.932 & 0.205 &  5.85 \\
        & counter  & 28.67 & 0.897 & 0.218 &  4.44 \\
        & flowers  & 21.06 & 0.553 & 0.405 & 11.37 \\
        & garden   & 26.65 & 0.817 & 0.191 & 13.70 \\
        & kitchen  & 30.36 & 0.911 & 0.149 &  5.08 \\
        & room     & 31.04 & 0.910 & 0.230 &  3.44 \\
        & stump    & 26.31 & 0.745 & 0.302 & 10.97 \\
        & treehill & 23.15 & 0.627 & 0.388 & 11.97 \\
        & \textbf{AVG} & \textbf{27.02} & \textbf{0.791} & \textbf{0.264} & \textbf{9.36} \\
        \hline
        \multirow{10}{*}{0.025}
        & bicycle  & 24.48 & 0.714 & 0.306 & 15.32 \\
        & bonsai   & 30.66 & 0.925 & 0.215 &  5.39 \\
        & counter  & 28.16 & 0.886 & 0.233 &  3.91 \\
        & flowers  & 20.82 & 0.537 & 0.423 &  9.83 \\
        & garden   & 26.42 & 0.805 & 0.207 & 11.66 \\
        & kitchen  & 29.77 & 0.902 & 0.163 &  4.35 \\
        & room     & 30.76 & 0.905 & 0.240 &  3.13 \\
        & stump    & 25.98 & 0.725 & 0.328 &  9.36 \\
        & treehill & 23.04 & 0.614 & 0.406 & 10.43 \\
        & \textbf{AVG} & \textbf{26.68} & \textbf{0.779} & \textbf{0.280} & \textbf{8.15} \\
        
        % \hline
        \bottomrule[1.2pt]
    \end{tabular}
    \end{adjustbox}
\label{table: our_mipnerf_hac}
\end{table}

\IEEEpubidadjcol

\begin{table}[h!]\small
    \centering
    \setlength{\abovecaptionskip}{0cm}
	\setlength{\belowcaptionskip}{0.2cm}
    \caption{Results of `ContextGS~\cite{wang2024contextgs} + Ours SALVQ' for each scene from Deep Blending dataset~\cite{hedman2018deep}.}
    \begin{adjustbox}{width=\linewidth}
    \begin{tabular}{m{3cm}<{\centering} |c|c c c c}
        % \hline
        \toprule[1.2pt]
        \textbf{\(\lambda\)} & \textbf{Scenes} & \textbf{PSNR}↑ & \textbf{SSIM}↑ & \textbf{LPIPS}↓ & \textbf{SIZE}↓ \\ 
        
        \hline
        \multirow{3}{*}{0.002}
        & drjohnson & 29.72 & 0.907 & 0.256 & 4.90 \\
        & playroom  & 30.73 & 0.910 & 0.265 & 3.58 \\
        & \textbf{AVG} & \textbf{30.23} & \textbf{0.908} & \textbf{0.260} & \textbf{4.24} \\
        \hline
        \multirow{3}{*}{0.004}
        & drjohnson & 29.65 & 0.905 & 0.262 & 3.81 \\
        & playroom  & 30.64 & 0.909 & 0.269 & 2.79 \\
        & \textbf{AVG} & \textbf{30.14} & \textbf{0.907} & \textbf{0.266} & \textbf{3.30} \\
        \hline
        \multirow{3}{*}{0.008}
        & drjohnson & 29.51 & 0.903 & 0.269 & 2.89 \\
        & playroom  & 30.45 & 0.905 & 0.278 & 2.20 \\
        & \textbf{AVG} & \textbf{29.98} & \textbf{0.904} & \textbf{0.274} & \textbf{2.54} \\
        \hline
        \multirow{3}{*}{0.015}
        & drjohnson & 29.24 & 0.898 & 0.282 & 2.24 \\
        & playroom  & 30.24 & 0.902 & 0.286 & 1.74 \\
        & \textbf{AVG} & \textbf{29.74} & \textbf{0.900} & \textbf{0.284} & \textbf{1.99} \\
        \hline
        \multirow{3}{*}{0.025}
        & drjohnson & 29.03 & 0.892 & 0.293 & 1.80 \\
        & playroom  & 29.99 & 0.899 & 0.295 & 1.40 \\
        & \textbf{AVG} & \textbf{29.51} & \textbf{0.895} & \textbf{0.294} & \textbf{1.60} \\
        \bottomrule[1.2pt]
    \end{tabular}
    \end{adjustbox}
\label{table: our_db_ContextGS}
\end{table}

\begin{table}[!ht]\small
    \centering
    \setlength{\abovecaptionskip}{0cm}
	\setlength{\belowcaptionskip}{0.2cm}
    \caption{Results of `ContextGS~\cite{wang2024contextgs} + Ours SALVQ' for each scene from Tank \& Temples dataset~\cite{knapitsch2017tanks}.}
    \begin{adjustbox}{width=\linewidth}
    \begin{tabular}{m{3cm}<{\centering} |c|c c c c}
        % \hline
        \toprule[1.2pt]
        \textbf{\(\lambda_r\)} & \textbf{Scenes} & \textbf{PSNR}↑ & \textbf{SSIM}↑ & \textbf{LPIPS}↓ & \textbf{SIZE}↓ \\ 
        
        \hline
        \multirow{3}{*}{0.002}
        & train & 22.74 & 0.823 & 0.211 & 7.58 \\
        & truck & 26.05 & 0.886 & 0.146 & 9.11 \\
        & \textbf{AVG} & \textbf{24.40} & \textbf{0.855} & \textbf{0.179} & \textbf{8.35} \\
        \hline
        \multirow{3}{*}{0.004}
        & train & 22.67 & 0.819 & 0.217 & 6.10 \\
        & truck & 26.02 & 0.885 & 0.150 & 7.53 \\
        & \textbf{AVG} & \textbf{24.35} & \textbf{0.852} & \textbf{0.184} & \textbf{6.81} \\
        \hline
        \multirow{3}{*}{0.008}
        & train & 22.48 & 0.812 & 0.228 & 4.88 \\
        & truck & 25.89 & 0.881 & 0.158 & 6.30 \\
        & \textbf{AVG} & \textbf{24.19} & \textbf{0.846} & \textbf{0.193} & \textbf{5.59} \\
        \hline
        \multirow{3}{*}{0.015}
        & train & 22.33 & 0.804 & 0.240 & 3.91 \\
        & truck & 25.76 & 0.878 & 0.165 & 4.84 \\
        & \textbf{AVG} & \textbf{24.05} & \textbf{0.841} & \textbf{0.202} & \textbf{4.37} \\
        \hline
        \multirow{3}{*}{0.025}
        & train & 22.07 & 0.793 & 0.253 & 3.11 \\
        & truck & 25.59 & 0.872 & 0.176 & 4.30 \\
        & \textbf{AVG} & \textbf{23.83} & \textbf{0.833} & \textbf{0.215} & \textbf{3.71} \\
       
        \bottomrule[1.2pt]
    \end{tabular}
    \end{adjustbox}
\label{table: our_tnt_ContextGS}
\end{table}

\begin{table}[!ht]\small
    \centering
    \setlength{\abovecaptionskip}{0cm}
	\setlength{\belowcaptionskip}{0.2cm}
    \caption{Results of `ContextGS~\cite{wang2024contextgs} + Ours SALVQ' for each scene from Mip-NeRF 360 dataset~\cite{barron2022mip}.}
    \begin{adjustbox}{width=\linewidth}
    \begin{tabular}{m{3cm}<{\centering} |c|c c c c}
        % \hline
        \toprule[1.2pt]
        \textbf{\(\lambda\)} & \textbf{Scenes} & \textbf{PSNR}↑ & \textbf{SSIM}↑ & \textbf{LPIPS}↓ & \textbf{SIZE}↓ \\ 
        \hline
        \multirow{10}{*}{0.002}
        & bicycle  & 25.03 & 0.739 & 0.265 & 25.24 \\
        & bonsai   & 32.85 & 0.948 & 0.182 &  8.24 \\
        & counter  & 29.57 & 0.915 & 0.191 &  7.42 \\
        & flowers  & 21.27 & 0.574 & 0.378 & 19.50 \\
        & garden   & 27.44 & 0.849 & 0.140 & 22.18 \\
        & kitchen  & 31.45 & 0.928 & 0.127 &  8.34 \\
        & room     & 31.85 & 0.925 & 0.202 &  5.51 \\
        & stump    & 26.64 & 0.762 & 0.266 & 16.78 \\
        & treehill & 23.27 & 0.645 & 0.350 & 19.75 \\
        & \textbf{AVG} & \textbf{27.71} & \textbf{0.809} & \textbf{0.233} & \textbf{14.77} \\
        \hline
        \multirow{10}{*}{0.004}
        & bicycle  & 25.05 & 0.738 & 0.270 & 20.64 \\
        & bonsai   & 32.57 & 0.946 & 0.187 &  6.99 \\
        & counter  & 29.34 & 0.911 & 0.198 &  5.97 \\
        & flowers  & 21.28 & 0.575 & 0.377 & 16.09 \\
        & garden   & 27.34 & 0.846 & 0.146 & 18.30 \\
        & kitchen  & 31.27 & 0.925 & 0.133 &  6.70 \\
        & room     & 31.71 & 0.923 & 0.208 &  4.24 \\
        & stump    & 26.64 & 0.763 & 0.269 & 14.32 \\
        & treehill & 23.36 & 0.646 & 0.354 & 15.79 \\
        & \textbf{AVG} & \textbf{27.62} & \textbf{0.808} & \textbf{0.238} & \textbf{12.12} \\
        \hline
        \multirow{10}{*}{0.008}
        & bicycle  & 25.01 & 0.736 & 0.277 & 17.42 \\
        & bonsai   & 32.17 & 0.942 & 0.193 &  5.58 \\
        & counter  & 29.13 & 0.907 & 0.205 &  4.68 \\
        & flowers  & 21.24 & 0.572 & 0.383 & 12.66 \\
        & garden   & 27.19 & 0.839 & 0.156 & 14.11 \\
        & kitchen  & 30.88 & 0.920 & 0.140 &  5.20 \\
        & room     & 31.50 & 0.919 & 0.216 &  3.43 \\
        & stump    & 26.66 & 0.765 & 0.271 & 11.60 \\
        & treehill & 23.26 & 0.646 & 0.356 & 12.98 \\
        & \textbf{AVG} & \textbf{27.45} & \textbf{0.805} & \textbf{0.244} & \textbf{9.74} \\
        \hline
        \multirow{10}{*}{0.015}
        & bicycle  & 25.02 & 0.732 & 0.286 & 13.66 \\
        & bonsai   & 31.81 & 0.938 & 0.200 &  4.54 \\
        & counter  & 28.75 & 0.901 & 0.215 &  3.76 \\
        & flowers  & 21.27 & 0.569 & 0.389 & 10.55 \\
        & garden   & 27.00 & 0.829 & 0.173 & 11.05 \\
        & kitchen  & 30.52 & 0.914 & 0.149 &  4.17 \\
        & room     & 31.20 & 0.914 & 0.227 &  2.74 \\
        & stump    & 26.60 & 0.761 & 0.280 &  9.53 \\
        & treehill & 23.27 & 0.643 & 0.366 & 10.40 \\
        & \textbf{AVG} & \textbf{27.27} & \textbf{0.800} & \textbf{0.254} & \textbf{7.82} \\
        \hline
        \multirow{10}{*}{0.025}
        & bicycle  & 24.82 & 0.722 & 0.301 & 11.42 \\
        & bonsai   & 31.39 & 0.933 & 0.209 &  4.16 \\
        & counter  & 28.50 & 0.894 & 0.225 &  3.17 \\
        & flowers  & 21.07 & 0.563 & 0.397 &  8.77 \\
        & garden   & 26.74 & 0.817 & 0.193 &  8.94 \\
        & kitchen  & 30.12 & 0.907 & 0.162 &  3.58 \\
        & room     & 30.96 & 0.907 & 0.241 &  2.32 \\
        & stump    & 26.55 & 0.757 & 0.291 &  8.29 \\
        & treehill & 23.16 & 0.638 & 0.377 &  8.48 \\
        & \textbf{AVG} & \textbf{27.03} & \textbf{0.793} & \textbf{0.266} & \textbf{6.57} \\
        
        % \hline
        \bottomrule[1.2pt]
    \end{tabular}
    \end{adjustbox}
\label{table: our_mipnerf_ContextGS}
\end{table}

\begin{table}[!ht]\small
    \centering
    \setlength{\abovecaptionskip}{0cm}
	\setlength{\belowcaptionskip}{0.2cm}
    \caption{Results of `HAC++~\cite{chen2025hac++} + Ours SALVQ' for each scene from Deep Blending dataset~\cite{hedman2018deep}.}
    \begin{adjustbox}{width=\linewidth}
    \begin{tabular}{m{3cm}<{\centering} |c|c c c c}
        % \hline
        \toprule[1.2pt]
        \textbf{\(\lambda\)} & \textbf{Scenes} & \textbf{PSNR}↑ & \textbf{SSIM}↑ & \textbf{LPIPS}↓ & \textbf{SIZE}↓ \\ 
        
        \hline
        \multirow{3}{*}{0.002}
        & drjohnson & 29.78 & 0.907 & 0.257 & 4.69 \\
        & playroom  & 30.91 & 0.911 & 0.261 & 3.32 \\
        & \textbf{AVG} & \textbf{30.34} & \textbf{0.909} & \textbf{0.260} & \textbf{4.01} \\
        \hline
        \multirow{3}{*}{0.004}
        & drjohnson & 29.71 & 0.905 & 0.264 & 3.36 \\
        & playroom  & 30.66 & 0.909 & 0.268 & 2.39 \\
        & \textbf{AVG} & \textbf{30.18} & \textbf{0.907} & \textbf{0.266} & \textbf{2.87} \\
        \hline
        \multirow{3}{*}{0.008}
        & drjohnson & 29.51 & 0.901 & 0.276 & 2.38 \\
        & playroom  & 30.49 & 0.901 & 0.282 & 1.70 \\
        & \textbf{AVG} & \textbf{30.00} & \textbf{0.901} & \textbf{0.279} & \textbf{2.03} \\
        \hline
        \multirow{3}{*}{0.015}
        & drjohnson & 29.21 & 0.894 & 0.290 & 1.70 \\
        & playroom  & 30.33 & 0.901 & 0.292 & 1.25 \\
        & \textbf{AVG} & \textbf{29.77} & \textbf{0.897} & \textbf{0.291} & \textbf{1.47} \\
        \hline
        \multirow{3}{*}{0.025}
        & drjohnson & 28.93 & 0.887 & 0.305 & 1.25 \\
        & playroom  & 29.98 & 0.896 & 0.303 & 1.01 \\
        & \textbf{AVG} & \textbf{29.46} & \textbf{0.892} & \textbf{0.304} & \textbf{1.13} \\
        \bottomrule[1.2pt]
    \end{tabular}
    \end{adjustbox}
\label{table: our_db_hacp}
\end{table}

\begin{table}[!ht]\small
    \centering
    \setlength{\abovecaptionskip}{0cm}
	\setlength{\belowcaptionskip}{0.2cm}
    \caption{Results of `HAC++~\cite{chen2025hac++} + Ours SALVQ' for each scene from Tank \& Temples dataset~\cite{knapitsch2017tanks}.}
    \begin{adjustbox}{width=\linewidth}
    \begin{tabular}{m{3cm}<{\centering} |c|c c c c}
        % \hline
        \toprule[1.2pt]
        \textbf{\(\lambda_r\)} & \textbf{Scenes} & \textbf{PSNR}↑ & \textbf{SSIM}↑ & \textbf{LPIPS}↓ & \textbf{SIZE}↓ \\ 
        
        \hline
        \multirow{3}{*}{0.002}
        & train & 22.61 & 0.820 & 0.213 & 5.95 \\
        & truck & 26.05 & 0.886 & 0.150 & 7.52 \\
        & \textbf{AVG} & \textbf{24.33} & \textbf{0.853} & \textbf{0.181} & \textbf{6.74} \\
        \hline
        \multirow{3}{*}{0.004}
        & train & 22.58 & 0.816 & 0.223 & 4.67 \\
        & truck & 25.94 & 0.882 & 0.157 & 5.77 \\
        & \textbf{AVG} & \textbf{24.26} & \textbf{0.849} & \textbf{0.190} & \textbf{5.22} \\
        \hline
        \multirow{3}{*}{0.008}
        & train & 22.33 & 0.804 & 0.240 & 3.49 \\
        & truck & 25.81 & 0.873 & 0.171 & 4.02 \\
        & \textbf{AVG} & \textbf{24.07} & \textbf{0.839} & \textbf{0.205} & \textbf{3.76} \\
        \hline
        \multirow{3}{*}{0.015}
        & train & 22.20 & 0.796 & 0.254 & 2.62 \\
        & truck & 25.59 & 0.869 & 0.187 & 2.83 \\
        & \textbf{AVG} & \textbf{23.89} & \textbf{0.832} & \textbf{0.221} & \textbf{2.73} \\
        \hline
        \multirow{3}{*}{0.025}
        & train & 22.09 & 0.784 & 0.274 & 1.98 \\
        & truck & 25.36 & 0.859 & 0.208 & 2.19 \\
        & \textbf{AVG} & \textbf{23.72} & \textbf{0.822} & \textbf{0.241} & \textbf{2.09} \\
       
        \bottomrule[1.2pt]
    \end{tabular}
    \end{adjustbox}
\label{table: our_tnt_hacp}
\end{table}

\begin{table}[!ht]\small
    \centering
    \setlength{\abovecaptionskip}{0cm}
	\setlength{\belowcaptionskip}{0.2cm}
    \caption{Results of `HAC++~\cite{chen2025hac++} + Ours SALVQ' for each scene from Mip-NeRF 360 dataset~\cite{barron2022mip}.}
    \begin{adjustbox}{width=\linewidth}
    \begin{tabular}{m{3cm}<{\centering} |c|c c c c}
        % \hline
        \toprule[1.2pt]
        \textbf{\(\lambda\)} & \textbf{Scenes} & \textbf{PSNR}↑ & \textbf{SSIM}↑ & \textbf{LPIPS}↓ & \textbf{SIZE}↓ \\ 
        \hline
        \multirow{10}{*}{0.002}
        & bicycle  & 25.16 & 0.738 & 0.276 & 17.50 \\
        & bonsai   & 32.86 & 0.947 & 0.185 &  6.91 \\
        & counter  & 29.64 & 0.915 & 0.193 &  6.24 \\
        & flowers  & 21.30 & 0.573 & 0.384 & 14.38 \\
        & garden   & 27.37 & 0.842 & 0.158 & 17.10 \\
        & kitchen  & 31.53 & 0.927 & 0.130 &  7.09 \\
        & room     & 31.99 & 0.924 & 0.207 &  4.56 \\
        & stump    & 26.67 & 0.762 & 0.275 & 12.42 \\
        & treehill & 23.30 & 0.643 & 0.366 & 14.52 \\
        & \textbf{AVG} & \textbf{27.76} & \textbf{0.808} & \textbf{0.241} & \textbf{11.19} \\
        \hline
        \multirow{10}{*}{0.004}
         & bicycle  & 25.07 & 0.731 & 0.291 & 12.74 \\
        & bonsai   & 32.55 & 0.944 & 0.190 &  5.32 \\
        & counter  & 29.42 & 0.910 & 0.202 &  4.75 \\
        & flowers  & 21.29 & 0.569 & 0.392 & 10.76 \\
        & garden   & 27.18 & 0.833 & 0.178 & 12.76 \\
        & kitchen  & 31.27 & 0.923 & 0.137 &  5.25 \\
        & room     & 31.77 & 0.920 & 0.217 &  3.45 \\
        & stump    & 26.59 & 0.759 & 0.285 &  9.36 \\
        & treehill & 23.33 & 0.639 & 0.379 & 10.42 \\
        & \textbf{AVG} & \textbf{27.61} & \textbf{0.803} & \textbf{0.252} & \textbf{8.31} \\
        \hline
        \multirow{10}{*}{0.008}
        & bicycle  & 24.97 & 0.721 & 0.306 &  9.20 \\
        & bonsai   & 32.22 & 0.939 & 0.199 &  3.97 \\
        & counter  & 29.14 & 0.903 & 0.213 &  3.52 \\
        & flowers  & 21.24 & 0.565 & 0.401 &  7.99 \\
        & garden   & 26.87 & 0.819 & 0.203 &  9.19 \\
        & kitchen  & 30.81 & 0.916 & 0.147 &  3.81 \\
        & room     & 31.57 & 0.915 & 0.229 &  2.58 \\
        & stump    & 26.56 & 0.754 & 0.297 &  6.95 \\
        & treehill & 23.25 & 0.633 & 0.395 &  7.51 \\
        & \textbf{AVG} & \textbf{27.40} & \textbf{0.796} & \textbf{0.266} & \textbf{6.08} \\
        \hline
        \multirow{10}{*}{0.015}
        & bicycle  & 24.80 & 0.707 & 0.326 &  6.40 \\
        & bonsai   & 31.78 & 0.936 & 0.206 &  3.03 \\
        & counter  & 28.83 & 0.895 & 0.229 &  2.61 \\
        & flowers  & 21.16 & 0.556 & 0.414 &  5.77 \\
        & garden   & 26.55 & 0.800 & 0.238 &  6.44 \\
        & kitchen  & 30.33 & 0.908 & 0.162 &  2.77 \\
        & room     & 31.25 & 0.908 & 0.244 &  1.88 \\
        & stump    & 26.46 & 0.745 & 0.316 &  5.01 \\
        & treehill & 23.21 & 0.623 & 0.414 &  5.34 \\
        & \textbf{AVG} & \textbf{27.15} & \textbf{0.786} & \textbf{0.283} & \textbf{4.36} \\
        \hline
        \multirow{10}{*}{0.025}
        & bicycle  & 24.65 & 0.690 & 0.347 & 4.74 \\
        & bonsai   & 31.27 & 0.929 & 0.217 & 2.33 \\
        & counter  & 28.42 & 0.882 & 0.250 & 1.97 \\
        & flowers  & 21.04 & 0.544 & 0.430 & 4.34 \\
        & garden   & 26.18 & 0.777 & 0.273 & 4.62 \\
        & kitchen  & 29.83 & 0.898 & 0.180 & 2.01 \\
        & room     & 30.94 & 0.900 & 0.262 & 1.47 \\
        & stump    & 26.26 & 0.732 & 0.336 & 3.75 \\
        & treehill & 23.12 & 0.610 & 0.435 & 3.88 \\
        & \textbf{AVG} & \textbf{26.86} & \textbf{0.773} & \textbf{0.303} & \textbf{3.24} \\
        
        % \hline
        \bottomrule[1.2pt]
    \end{tabular}
    \end{adjustbox}
\label{table: our_mipnerf_hacp}
\end{table}

\end{document}